\pdfoutput=1
\documentclass[11pt]{article}

\usepackage[]{acl}

\usepackage{times}
\usepackage{latexsym}

\usepackage[T1]{fontenc}
\usepackage[utf8]{inputenc}

\usepackage{microtype}

\usepackage{subcaption}
\usepackage{appendix}
\usepackage{xcolor}
\usepackage{graphicx}

\title{TIDE: Textual Identity Detection for Evaluating and Augmenting Classification and Language Models}

\author{Emmanuel Klu\thanks{These authors contributed equally to this work.} \and Sameer Sethi\footnotemark[1] \\
  Google Research \\
  \texttt{eklu@google.com}, \texttt{sethis@google.com} \\}

\begin{document}
\maketitle
\begin{abstract}
Machine learning models can perpetuate unintended biases from unfair and imbalanced datasets. Evaluating and debiasing these datasets and models is especially hard in text datasets where sensitive attributes such as race, gender, and sexual orientation may not be available. When these models are deployed into society, they can lead to unfair outcomes for historically underrepresented groups. In this paper, we present a dataset coupled with an approach to improve text fairness in classifiers and language models. We create a new, more comprehensive identity lexicon, TIDAL, which includes 15,123 identity terms and associated sense context across three demographic categories. We leverage TIDAL to develop an identity annotation and augmentation tool that can be used to improve the availability of identity context and the effectiveness of ML fairness techniques. We evaluate our approaches using human contributors, and additionally run experiments focused on dataset and model debiasing. Results show our assistive annotation technique improves the reliability and velocity of human-in-the-loop processes. Our dataset and methods uncover more disparities during evaluation, and also produce more fair models during remediation. These approaches provide a practical path forward for scaling classifier and generative model fairness in real-world settings. The code and dataset are available at \url{https://github.com/google-research/google-research/tree/master/tide_nlp}.
\end{abstract}

\section{Introduction}

The growing adoption of machine learning across a variety of applications have reignited concerns about unfair and unintended bias in models. Bias can be introduced throughout the development workflow, for example during problem framing, data sampling and preparation, and even through training algorithm choices \citep{shah-etal-2020-predictive,saleiro2018aequitas}. When models contain biases, they can play an active role in perpetuating societal inequities and unfair outcomes for underrepresented groups \citep{sweeney2013discrimination,abid2021persistent}.

Algorithmic fairness is a rapidly growing field of research with a wide range of definitions, techniques and toolkits available. Fairness is anchored in understanding and mitigating model performance disparities across sensitive and protected attributes. Popular toolkits such as AI Fairness 360 \citep{bellamy2018ai}, Fairlearn \citep{bird2020fairlearn}, and the Responsible AI toolkit in TensorFlow \citep{tensorflow2015-whitepaper}, all assume these attributes are readily available in datasets. In many real-world datasets, attributes are either not available or not reliable. This is due to a myriad of issues like privacy and safety laws around protected attributes, human annotation cost and reliability, and inconsistent taxonomy and attribute coverage \citep{andrus-spitzer-2021}. 

Attempts to address this problem involve techniques to extract attributes from text, through human or computational means. A common one is to create an adhoc list of ``identity terms'' \citep{dixon2018measuringmitigating} for token matching. However this approach is limited due to the polysemy of words (e.g. ``black'' as a color or race), scalability of token matching techniques, and a lack of important contextual information about the terms \citep{blodgett-etal-2020-language}. Connotation is one such example of missing context: a non-literal meaning of a word informed by one's beliefs and prejudices about its typical usage (e.g. ``undocumented workers'' and ``illegal aliens'' have the same lexical denotation but different connotations) \citep {carpuat-2015-connotation,pragmatics2007,webson-etal-2020-undocumented}. 

Our research goal is to first explore techniques that can improve availability and reliability of identity term annotations by providing context for disambiguation. A second goal is to leverage these annotations to adapt existing fairness techniques in ways that scale for use in real-wold text datasets and throughout the development workflow.

\subsection{Related Work}

\subsubsection{Availability of identity labels.} \citeauthor{gupta2018proxy,jung2022learning} propose methods to leverage proxy attributes in the absence of identity labels, however \citeauthor{tschantz2022,mcloughney2023} show proxies could be a source of bias and discrimination. When labels exist but are noisy or unreliable, \citeauthor{celis2021fair} explore techniques to achieve fairness under uncertainty. \citeauthor{lahoti2020fairness} attempt to remove the need for identity labels altogether. Our work follows \citet{andrus-villeneuve-2022}, focusing on addressing the issue earlier in the pipeline by taking a human-in-the-loop approach. We deploy assistive techniques for acquiring high quality annotations from humans faster.

\subsubsection{Identity lexicon.} \citep{eckle-kohler-etal-2012-uby} show the need for a standardized lexicon, while \citep{allaway-mckeown-2021-unified} extend one with contextual dimensions including sentiment and emotional association. Our approach is most closely related to \citep{smith-etal-2022-im} who create a similar identity lexicon. We focus on creating an extensible schema that enables multilingual support, and enabling fairness use cases by capturing additional context and increasing the depth of coverage across groups

\subsubsection{Identity entity recognition.}  Sense disambiguation \citep{pal2015word} has been used to address polysemy, with recent advances in knowledge-based techniques \citep{agirre-etal-2014-random}. On the other hand \citep{spacy2,nltkbook} use syntactic and NLP techniques to detect canonical entities like ``person'', which is too coarse. Our work merges both techniques to build a reusable annotation tool. We specialize in identity detection and optimize for fairness workflows, and additionally adapt for counterfactual generation.

\subsubsection{Effectiveness of fairness techniques.} \citep{dixon2018measuringmitigating} use a keyword list to source new organic data for debiasing datasets, while \citep{wadhwa2022fairness} generate counterfactuals using existing datasets as the seed. Our experiments aim to scale up both fairness techniques for use throughout the entire ML workflow. We also leverage identity taxonomy instead of terms to uncover previously missed bias in classifiers and generative models alike.

\subsection{Contributions} 
Our key contributions are summarized below:
\begin{itemize}
    \item Textual Identity Detection and Augmentation Lexicon (TIDAL): to the best of our knowledge TIDAL is the largest identity lexical dataset with comprehensive coverage of groups and associated sense context, using a methodology and schema that supports multiple languages.
    \item A specialized identity annotation tool built with the lexicon and optimized for multiple fairness workflows.
    \item An assistive technique for human annotation that improves time, cost and reliability of acquiring identity labels.
    \item Updated fairness techniques that improve coverage of bias detection and result in more effective remediation of datasets and models.
\end{itemize}

\subsection{Preliminaries}
\subsubsection{Datasets.} We use the CivilComments dataset \citep{borkan2019nuanced} for most experiments conducted, relying on its human-annotated identity labels as ground truth. We use the C4 dataset \citep{c4:2020} as a control.

\subsubsection{Data Augmentation.} We generate synthetic datasets using sentence templates from HolisticBias \citep{smith-etal-2022-im} and UnintendedBias \citep{dixon2018measuringmitigating}. We additionally generate counterfactuals \citep{wadhwa2022fairness} for robustness.

\subsubsection{Models.} For generative tasks we use BlenderBot \citep{roller-etal-2021-recipes}. For classification we train toxicity models on CivilComments, and additionally use counterfactual logit pairing (CLP) for remediation.

\subsubsection{Dataset and model evaluation metrics.} We use slice analysis and deficits to understand class balance in datasets and models \citep{dixon2018measuringmitigating}. We measure model performance using F1, area-under-curve (AUC), and counterfactual flips \citep{garg2019counterfactual} for classifiers, and token likelihood \citep{smith-etal-2022-im} for generative models.

\subsubsection{Inter-annotator reliability (IAR).} Following \citep{lacy2015issues}, we use simple percent agreement, Krippendorff's alpha \citep{krippendorff1970estimating} and Gwet's AC1 \citep{gwet2014handbook} to measure the degree of agreement on annotations between human annotators. While Krippendorff's alpha penalizes for data scarcity, Gwet's AC1 corrects for the probability that the annotators agree by chance - both cases are likely given our data distribution and task complexity.

\subsubsection{Identity terms and sense context.} Multiple descriptors are used throughout the literature to describe words, utterances or context associated with identity, such as ``sensitive attributes'', ``sensitive features'', ``group labels'', ``protected attributes'' or ``identity terms'' \citep{garg2019counterfactual,dixon2018measuringmitigating}. In our work we use ``identity terms'' for the lexicon that appears in text, and ``sense context'', for the structured contextual data associated with senses of identity terms.

\section{Methodology}
\subsection{TIDAL dataset}
The TIDAL dataset consists of lexical entries and their related forms (e.g. black, gay, trans, hindus) that are associated with identity groups. Each head and related form is associated with grammatical properties (e.g. part-of-speech, grammatical gender) and context (or ``sense'') entries (e.g. identity groups/subgroups, connotation). Although we develop a lexicon, schema and methodology that works for multiple languages, we will focus on English in this paper. In total TIDAL has 1,419 English language head-form identity lexical entries, with over 13,709 related lexical forms and 15,270 context/sense entries.

\begin{figure}[htbp]
\centering
\includegraphics[width=1\columnwidth]{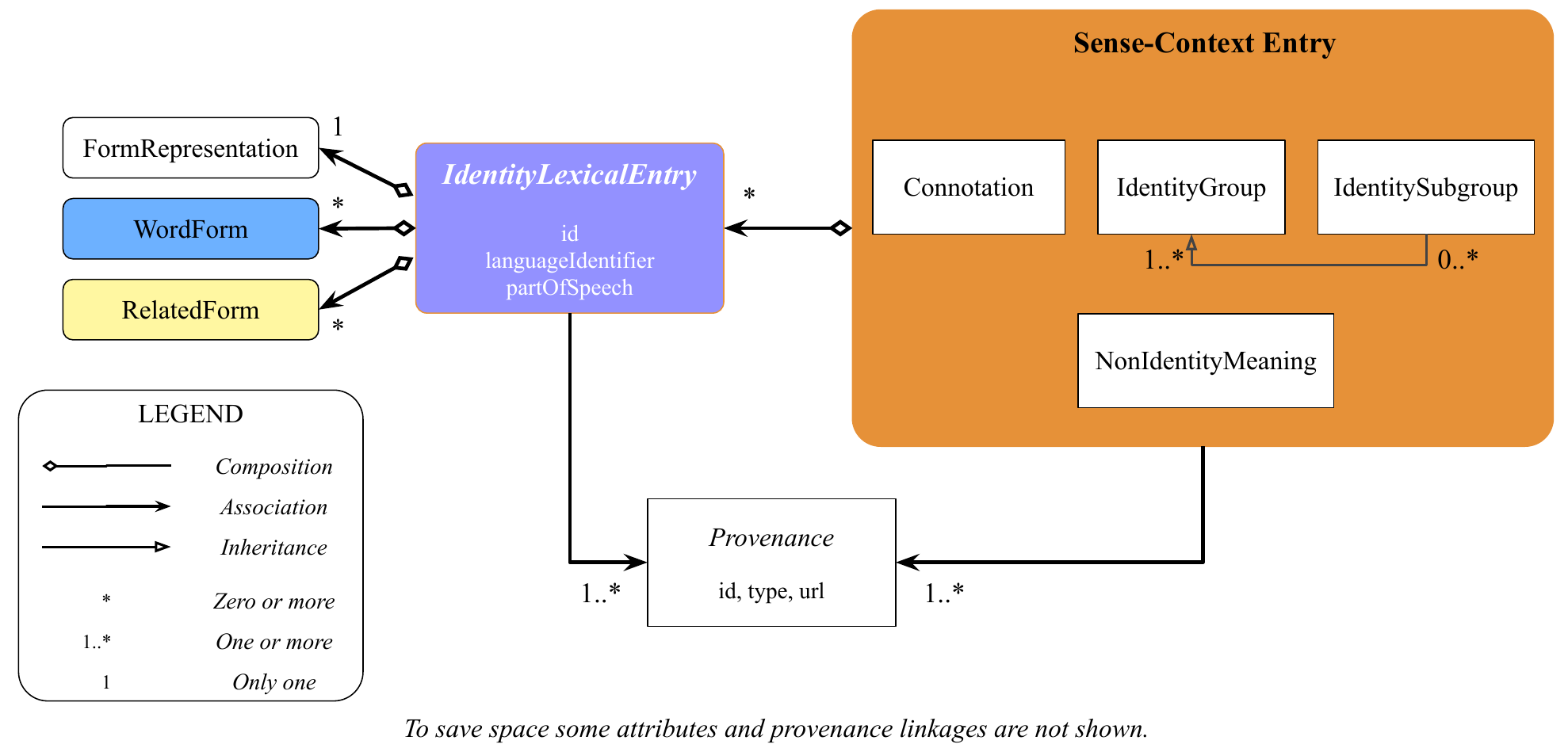}
\caption{TIDAL: Conceptual model}
\label{fig:tidal_conceptual_model}
\end{figure}

\begin{figure}[htbp]
\centering
\includegraphics[width=1\columnwidth]{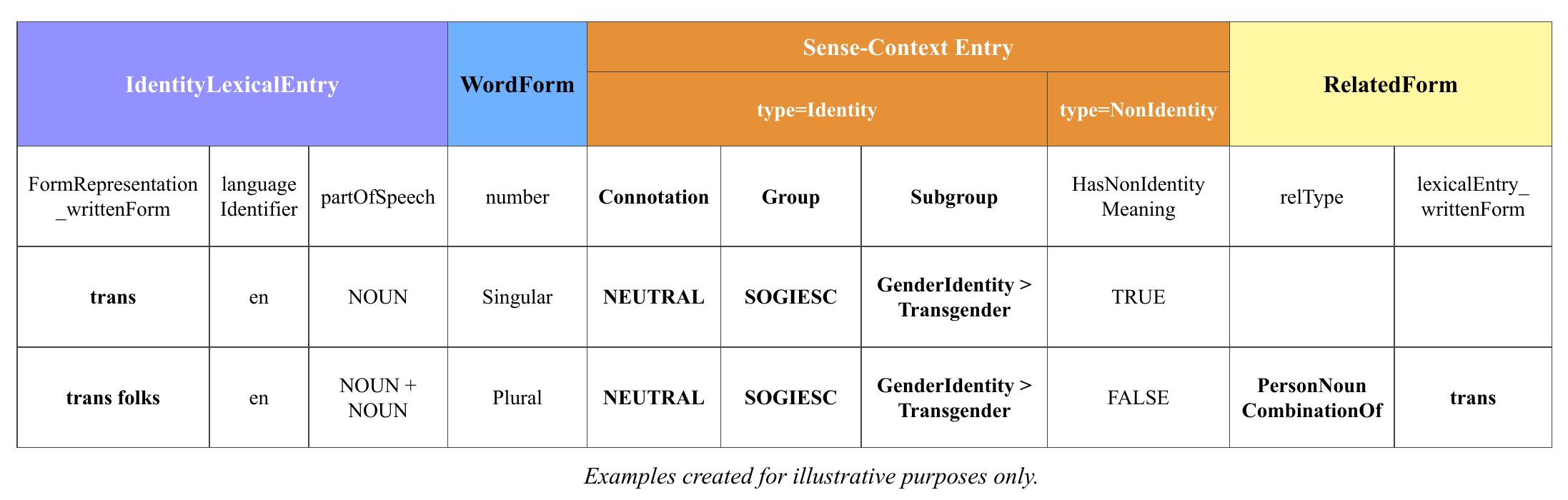}
\caption{TIDAL: Example, flattened tabular format.}
\label{fig:tidal_flattened_example}
\end{figure}

\subsubsection{Schema.}
Figure~\ref{fig:tidal_conceptual_model} shows the conceptual model of the TIDAL schema and Figure~\ref{fig:tidal_flattened_example} shows a flattened tabular example of TIDAL data. We create an adapted UBY-LMF schema \citep{eckle-kohler-etal-2012-uby} which is based on the Lexical Markup Framework (LMF) standard \citep{ISO2461372:online} for representing NLP lexicons. 

Our paper focuses on the following identity groups (IdentityGroup): race, nationality or ethnicity (RNE), sexual orientation, gender identity, gender expression and sex characteristics (SOGIESC) and Religion. We choose RNE as a collective category to be more inclusive since their constituent concepts of race, ancestry, nationality and ethnicity are inconsistent and sometimes redundant across cultures \citep{morning2008ethnic}. We choose SOGIESC for similar reasons, instead of Gender Identity and Sexual Orientation, LGBTI or SOGI \citep{trithart2021}. Although multiple dimensions of connotation like social value, politeness or emotional association have been proposed in prior lexical work \citep{allaway-mckeown-2021-unified}, our scope is limited to NEUTRAL and PEJORATIVE connotations. PEJORATIVE implies a term can be used to demean or disparage a group of people.

Table~\ref{table:lexicon_comparison} shows a comparative analysis of TIDAL with known similar sources such as UnintendedBias \citep{dixon2018measuringmitigating} used by Perspective API \footnote{\url{https://perspectiveapi.com/}}, and HolisticBias \citep{smith-etal-2022-im}. Additional details of our data distribution can be found in Appendix \ref{appendix_tidal_distribution}.

\subsubsection{Sourcing.} We source the seed set of identity terms for our lexicon from the following public sources:
\begin{itemize}
    \item \textbf{UNdata} \citep{united2003ethnicity}: ``Population by national and/or ethnic group'' and ``Population by religion'' tables from UNData are used to create RNE and Religion seed sets, respectively.
    \item \textbf{CAMEO} \citep{Gerner2002ConflictAM}:  We utilize the CAMEO coding framework, which contains approximately 1,500 religions and 650 ethnic groups.
    \item \textbf{GLAAD}: We leverage GLAAD glossary of LGBTQ and transgender terms \citep{GLAADGlossary:online} for SOGIESC seed sets.
    \item \textbf{HRC}: We use HRC glossary of words and meanings \citep{HRCGlossary:online} for SOGIESC seed sets.
    \item \textbf{Wikipedia}: We leverage demonyms and adjectivals \citep{WikipediaDemonyms:online} list for RNE seed sets.
\end{itemize}

Appendix \ref{appendix_post_processing_seed_set} provides additional details on seed set data processing.

\subsubsection{Curation.} We expand the seed terms to their grammatical and morphological variants using linguistic experts and rule-based lexical expansion tools. Each resulting term is treated as a new lexical entry with reference to the head. Next we curate multiple pools of data contributors to corroborate, correct and expand our data. We leverage a human annotation platform to curate a diverse pool of linguistic experts and create tasks reflecting the following phases: 
\begin{enumerate}
    \item \textbf{Expansion}: expand seed terms to grammatical variants, common misspellings and person noun combinations.
    \item \textbf{Contextualization}: research and associate all possible context for seed terms and expansions, including connotation and identity groups.
    \item \textbf{Disambiguation}: research and associate context that can help distinguish identity and prevalent non-identity usage of the terms.
\end{enumerate}

Contributors research public sources (such as dictionaries, encyclopedias, and other lexical sources) for unstructured context for identity terms. They also provide citations for the sources they use, their own beliefs about missing context or usage of a term not available in sources. Finally, we anonymize contributor personally-identifiable information before aggregating the assertions and ingesting the data into the lexicon database.

\begin{table}[htbp]
\resizebox{.95\columnwidth}{!}{\begin{tabular}{|p{1in}|c|c|c|}\hline
     & HolisticBias & UnintendedBias & \textbf{TIDAL} \\
    \hline
    Supported Identity Groups & 14 & N/A & \textbf{3} \\
    \hline
    Head terms / lexical entries & 594 & 50 & \textbf{1565} \\
    \hline
    Variants and expansions & - & - & \textbf{14148} \\
    \hline
    Includes connotation context & No & No & \textbf{Yes} \\
    \hline
    Includes identity groups/subgroups & Yes & No & \textbf{Yes} \\
    \hline
    Includes non-identity context & No & No & \textbf{Yes} \\
    \hline
\end{tabular}}
\caption{Comparison of TIDAL to other lexicons datasets.}
\label{table:lexicon_comparison}
\end{table}

\begin{figure}[htbp]
\centering
\includegraphics[width=1\columnwidth]{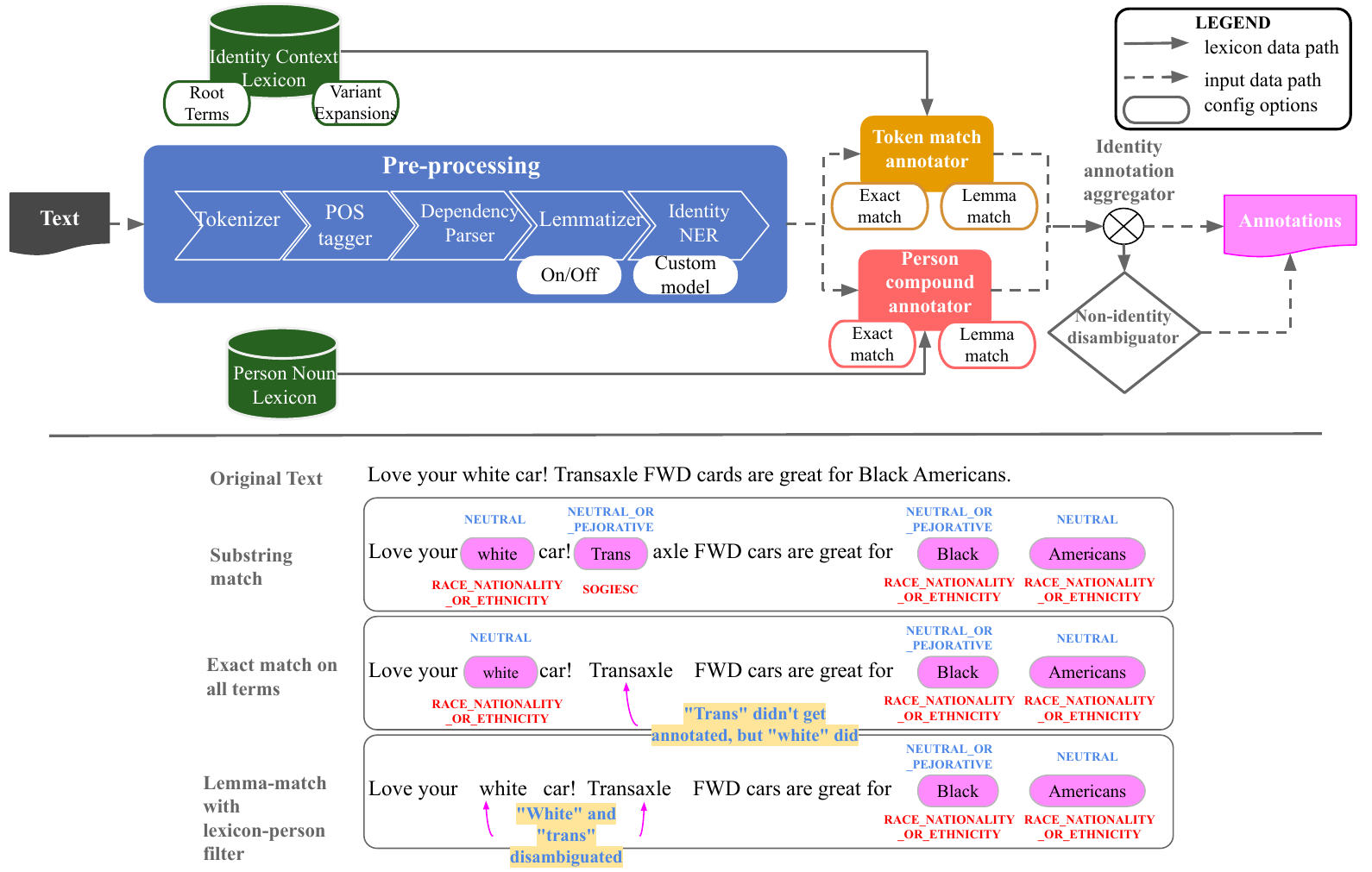} % Reduce the figure size so that it is slightly narrower than the column. Don't use precise values for figure width.This setup will avoid overfull boxes.
\caption{Data flow and system components of the annotation tool, with examples.}
\label{fig:annotation_flow}
\end{figure}

\subsection{Identity Annotation Tool}
To scale the acquisition of identity labels, we build a configurable multi-label multi-class annotation tool that leverages our identity lexicon and lexical properties to label identity terms found in text.

\subsubsection{Annotator components.} We first preprocess text using spaCy \citep{spacy2} to tokenize and tag with part-of-speech labels, the dependency tree and morphological properties. We then match tokens with terms in the lexicon, using lemmas and variants. We disambiguate non-identity usage of terms with person-noun detection using i) a lexicon of person nouns from Wiktionary \citep{WikiTermsOfAddress:online} and ii) the NLTK \citep{nltkbook} wordnet module to compare similarity with person identifiers like ``person'' and ``people'' and non-person identifiers like ``object'' and ``thing''. Additionally, spaCy linguistic features \citep{spacy2} is used for person-nouns detection using named entities like ``PERSON'', ``NORP'', and ``GPE''. To disambiguate a potential identity term we use the dependency tree (with support for conjunctions) and part-of-speech tags to include tokens that modify person-nouns and exclude tokens that modify non-person nouns. Finally, we train a custom spaCy NER model. The output of the annotator includes identity groups, subgroups, connotation and possible non-identity usage. Figure~\ref{fig:annotation_flow} shows the annotation flow and example output. Additional design details are specified in Appendix \ref{appendix_annotation_details}.

\section{Acquiring Identity Context at Scale}
\subsection{Annotation Tool Performance}
We measure the performance of our annotation techniques against human annotations available in the CivilComments dataset, and additionally validate performance consistency using the C4 dataset as a control. Our goal is to understand the effectiveness of techniques for a variety of downstream tasks, and whether performance can generalize to new datasets.

\subsubsection{Annotation techniques.} We implement substring matching as the baseline technique and configure multiple annotator variants using tokenizers: i) tokenize and match any occurrence in the lexicon, including all term forms and expansions; ii) tokenize and match occurrence of head terms only; iii) a variation of ii) that additionally disambiguates using a person-term lexicon; and iv) a variation of iii) that uses similarity-to-person-term disambiguation. We finally configure the custom NER model as a standalone annotator variant. Across all techniques, only annotations matching lexical entries in the dataset are considered valid. Figure~\ref{fig:annotation_flow} shows examples of annotation output.

\subsubsection{F1 scores.} 
\begin{figure}[htbp]
\centering
\includegraphics[width=0.9\columnwidth]{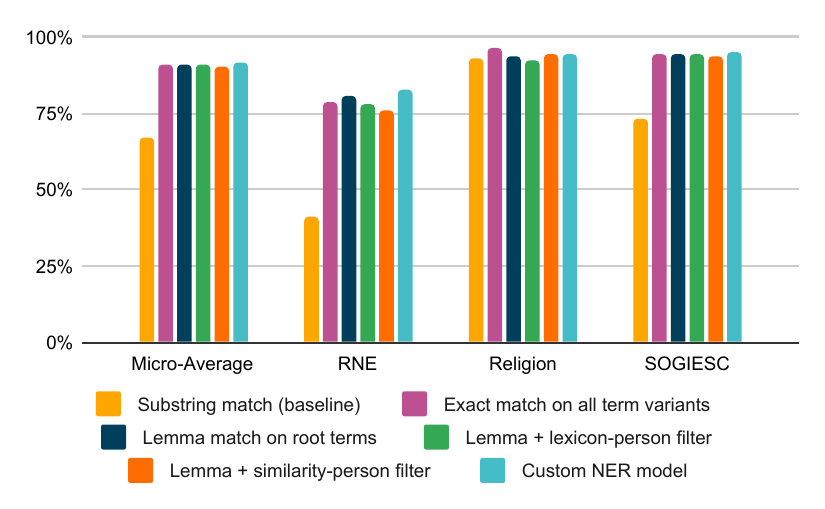} % Reduce the figure size so that it is slightly narrower than the column. Don't use precise values for figure width.This setup will avoid overfull boxes.
\caption{Multi-class F1 scores for the identity annotation tool on CivilComments.}
\label{fig:annotation_performance}
\end{figure}

All techniques outperform substring matching, with the custom NER model achieving the highest score of 91.92\%, followed by lemma and exact matching (91.13\%, 91.11\%) in Figure~\ref{fig:annotation_performance}. Disambiguation filters result in increased false negatives that impact overall performance. RNE has the lowest performance trend among subgroups while Religion has the most similar performance across techniques. Additional performance details are provided in Appendix \ref{appendix_annotation_results}.

\subsection{Human Annotation Impact}
We assess the impact of assistive annotation in human annotation workflows used to acquire identity labels. In addition to time and cost improvements we seek to understand the quality and consistency of human annotations, including potential new biases.

\subsubsection{Methodology.} We sample 337 examples from the CivilComment dataset annotated in the previous experiment. This example dataset is balanced across groups and highlights the performance differences between annotator variants. We present these examples in a human computation task for contributors to first identify tokens associated with identity and then provide an appropriate IdentityGroup label (RNE, Religion or SOGIESC). From a pool of more than 1,000 human annotators, at least 5 annotators review each example. We run three variations of this human annotation task, i) the first with an example-only dataset as the baseline, and the others with assistive annotations: ii) using a token-matching annotator without disambiguation, and iii) using a token-matching annotator with disambiguation. We also request an optional satisfaction survey for each task where the human annotators are asked to rate ``Ease of Job'' and ``Pay''. We run the same set of experiments on the C4 dataset as a control. Detailed human annotation job design and guidelines can be found in Appendix \ref{appendix_hcomp_exp1_details}.

\subsubsection{Inter-annotator reliability (IAR).} 
\begin{figure}[htbp]
\centering
\includegraphics[width=0.9\columnwidth]{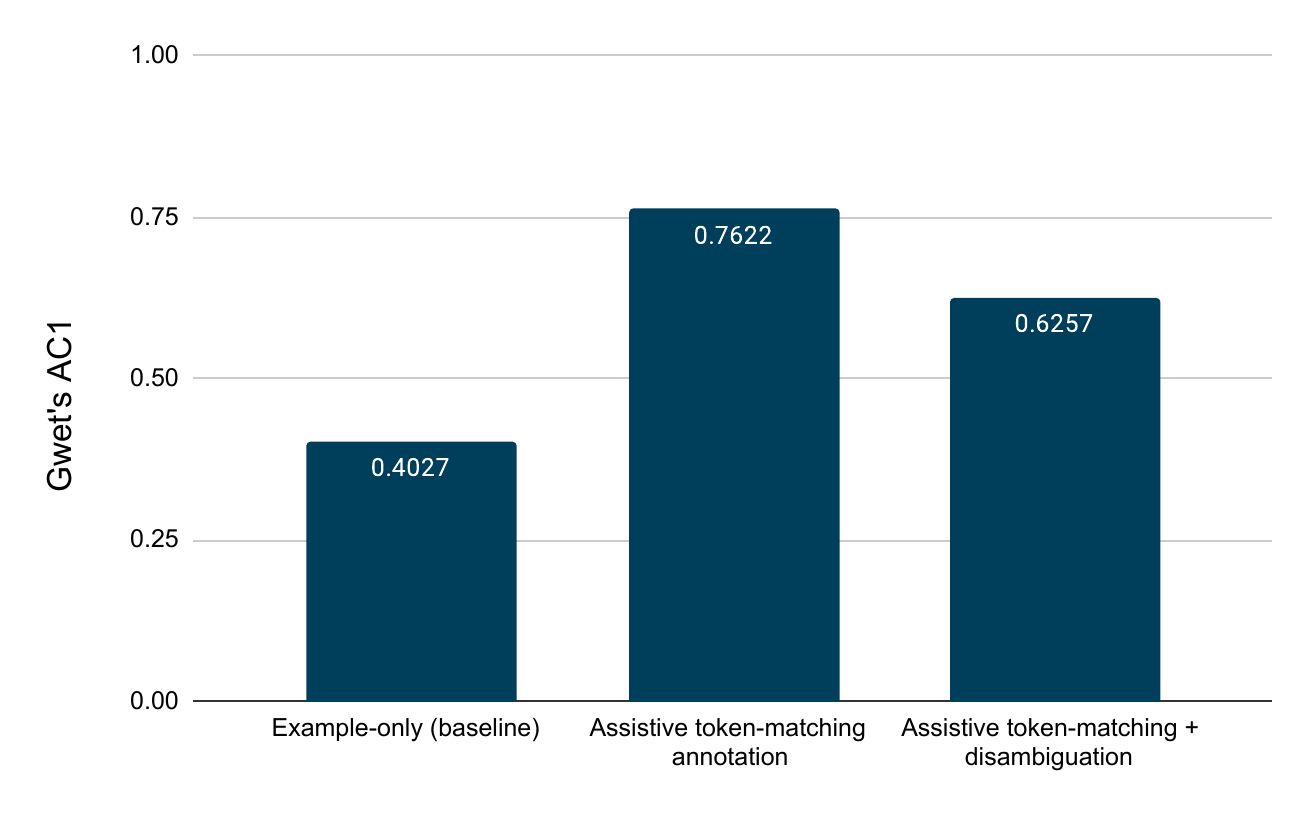} % Reduce the figure size so that it is slightly narrower than the column. Don't use precise values for figure width.This setup will avoid overfull boxes.
\caption{IAR (Gwet's AC1) for human annotations: identity labeling on CivilComments.}
\label{fig:hcomp_exp1_iar_cc}
\end{figure}

Assistive annotations consistently improve the reliability of human annotations as seen in Figure~\ref{fig:hcomp_exp1_iar_cc}. Token-matching achieves an Gwet's AC1 score of 0.7622,  representing a 89.27\% increase over the baseline, while additional disambiguation results in a score of 0.6257, a 55.37\% increase. Our analysis finds similar improvement trends in percent agreement and Krippendorff's Alpha metrics. Additional results are available in Appendix \ref{appendix_hcomp_exp1_results}.

\subsubsection{F1 scores.} 
\begin{figure}[htbp]
\centering
\includegraphics[width=0.9\columnwidth]{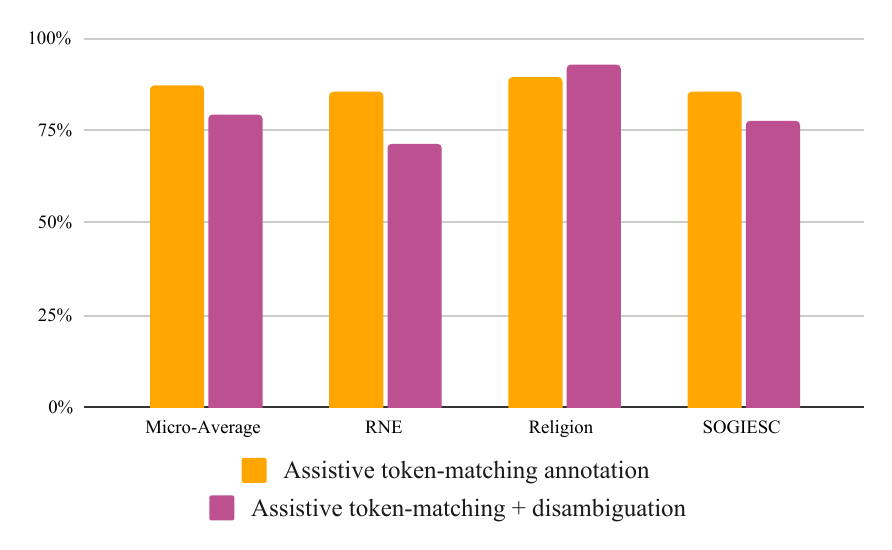} % Reduce the figure size so that it is slightly narrower than the column. Don't use precise values for figure width.This setup will avoid overfull boxes.
\caption{Multi-class F1 scores for human annotations: identity labeling on CivilComments.}
\label{fig:hcomp_exp1_f1_cc}
\end{figure}

Since IAR doesn't provide a per-class understanding of agreement and quality, we use micro-average F1 scores to understand performance across groups. We use the output of the baseline annotation task (example-only) as ground truth for this comparison. Token-matching achieves the highest overall score of 87.38\%, while additional disambiguation performs better only for Religion, seen in Figure~\ref{fig:hcomp_exp1_f1_cc}. Further analysis reveals tradeoffs between false positives and false negatives across the two annotation techniques. More details are in Appendix \ref{appendix_hcomp_exp1_results}. 

\subsubsection{Velocity, cost and satisfaction scores.} We use the interquartile mean (IQM) of time taken for a human annotator to complete the tasks as a proxy for completion velocity. To understand cost, we count the total number judgements required to meet the agreement threshold of 0.7. Lastly, the results from a task satisfaction survey inform task completion difficulty. Token-matching performs the best on velocity, taking 44.8\% less time than the baseline. Both assistive annotations tasks have similar costs (24-27\% better compared to the baseline). While we receive no data on satisfaction for token-matching, contributors find assistive annotations with disambiguation makes tasks 84.4\% easier to perform and result in 43.4\% better pay to the baseline task. Table~\ref{table:hcomp_exp1_velocity} provides detailed per task scores.

\begin{table}[htbp]
\resizebox{.95\columnwidth}{!}{\begin{tabular}{|p{1in}|c|c|c|c|}\hline
     & \textbf{Velocity} & \textbf{Cost} & \textbf{Ease of Job}  & \textbf{Pay} \\
    \hline
     & Judgement Time (s) & Total Judgements  & Scale: 1-5 & Scale: 1-5 \\
    \hline
    Example-only (baseline) & 82.5 & 2623 & 2.25 & 3 \\
    \hline
    Assistive annotations using token-matching & 45.5 & 1981 & - & - \\
    \hline
    Assistive annotations with disambiguation & 64 & 1905 & 4.15 & 4.3 \\
    \hline
\end{tabular}}
\caption{Velocity, cost and satisfaction results from human annotation tasks for identity labels}
\label{table:hcomp_exp1_velocity}
\end{table}

\section{Fairness Applications}
Our experiments in this section explore opportunities to leverage our lexicon and annotation tool at various points in the ML fairness workflow, from data labeling to model training. We modify and augment existing techniques from the literature in ways that are only enabled by our work. Our goal is to improve overall effectiveness of fairness interventions and demonstrate that it can be done at scale.

\subsection{Assistive Context for Ground Truth Labeling}
\begin{figure}[htbp]
\centering
\includegraphics[width=0.9\columnwidth]{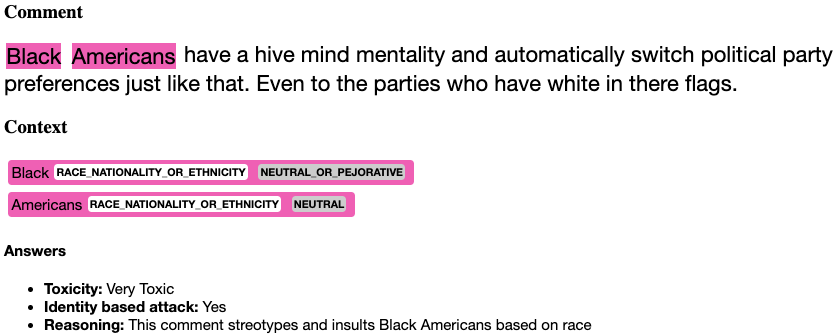}
\caption{Example of identity context annotation in HCOMP toxicity labeling task.}
\label{fig:hcomp_exp2_task_example}
\end{figure}

We explore data collection interventions by replicating the toxicity labeling human annotation task\footnote{\url{https://github.com/conversationai/conversationai.github.io}} for the Perspective API. Figure~\ref{fig:hcomp_exp2_task_example} shows an example of the assistive annotations we provide during human computation to understand the impact of context on annotation quality.

\subsubsection{Methodology.} We modify their human computation setup by excluding all sub-attributes except ``Identity based attack'', which we show only when the toxicity question is answered with ``VERY TOXIC'', ``TOXIC'' or ``NOT SURE''. We sample 298 examples from the CivilComment dataset annotated in the previous experiment, only including examples where our annotations are an exact match with provided ground truth labels. This example dataset is balanced across groups and is representative of the performance differences between annotator variants. We run three variations of the human evaluation task, i) the first with an example-only dataset as the baseline, and the others with assistive identity context: ii) providing ``IdentityGroup'' annotations, and iii) providing ``IdentityGroup'' and ``Connotation'' annotations. From a pool of more than 1,300 human annotators, at least 10 annotators review each example. Detailed human annotation job design and guidelines are given in Appendix \ref{appendix_hcomp_exp2_toxicity_details}.

\subsubsection{Inter-annotator reliability (IAR).} 
\begin{figure}[htbp]
\centering
\includegraphics[width=0.9\columnwidth]{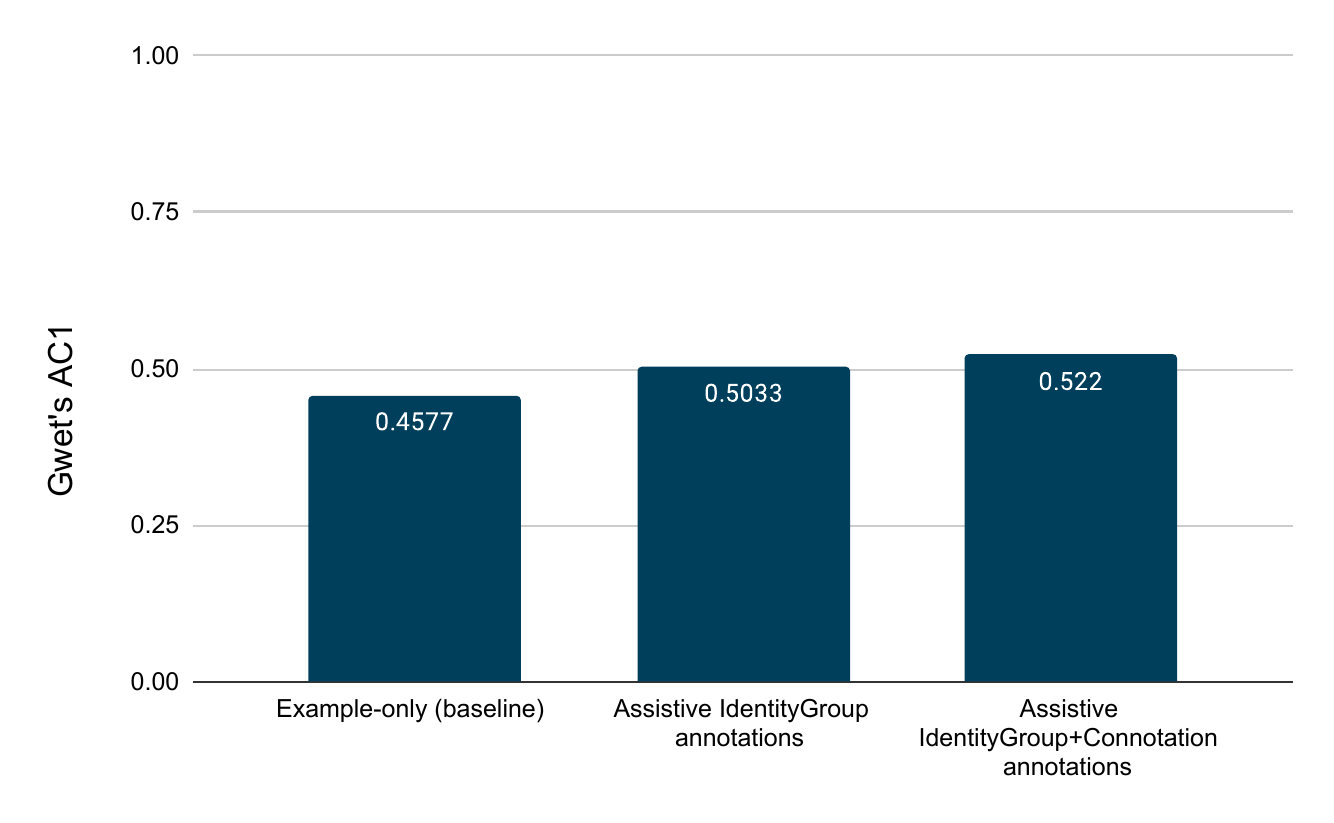}
\caption{IAR for human annotations: toxicity labeling on CivilComments.}
\label{fig:hcomp_exp2_iar}
\end{figure}

Assistive annotations consistently improve the reliability of human annotations as seen in Figure~\ref{fig:hcomp_exp2_iar}. IdentityGroup+Connotation annotations achieve the highest AC1 score, seeing an 14.04\% increase over the baseline, IdentityGroup annotations achieve an 9.96\% increase over baseline. Krippendorff's Alpha scores have the lowest trend due to class imbalance - 85\% of labels are toxic. Our agreement performance is consistent with prior work (\citep{ross2017measuring} and \citep{wulczyn2017ex}), given the subjective nature of toxicity labeling. Additional results are in Appendix \ref{appendix_hcomp_exp2_toxicity_results}.

\subsection{Counterfactual Logit Pairing}
We replicate the experimental setting from the counterfactual logit pairing (CLP) guide\footnote{\url{https://www.tensorflow.org/responsible_ai/model_remediation/counterfactual/guide/counterfactual_keras}}, and introduce additional counterfactual techniques enabled by our work to evaluate and mitigate classifier bias.

\subsubsection{Counterfactual techniques.} We establish a baseline with token ablation using their keyword list. We implement two additional techniques: i) token ablation using subgroup annotations instead of keywords and ii) token replacement using least similar counterfactuals. We train CLP-remediated models for each technique and evaluate flips on the baseline test set. Additional details in Appendix \ref{appendix_counterfactuals}.

\subsubsection{Counterfactual flip rates.} The counterfactual flip rate diff metric measures the difference between the flip rate for a counterfactual model and that of the base model on the baseline counterfactual dataset. Results show that using annotations for ablation instead of a keyword list increases the coverage of terms, leading to consistently fewer counterfactual flips in Table~\ref{table:counterfactual_flip_rates}. We also observe that the counterfactual ablation technique performs better than replacement since ablation creates only one counterfactual compared to multiple generated with replacement technique. Mitigating using counterfactual replacements requires generating multiple counterfactuals for better chances of success, which we'll observe in the next section. The CLP library also only supports generating one counterfactual which limits the coverage of counterfactual evaluation and remediation.

\begin{table}[htbp]
\resizebox{.95\columnwidth}{!}{\begin{tabular}{|p{1in}|c|c|c|c|}\hline
     & \textbf{Overall} & \textbf{Black} & \textbf{Homosexual} & \textbf{GenderIdentity} \\
    \hline
    Keyword ablation (baseline) & 0.37\% & 0.27\% & -0.30\% & 0.32\% \\
    \hline
    Annotation ablation & 0.08\% & -0.09\% & -0.74\% & 0.00\% \\
    \hline
    Annotation replacement & 0.34\% & 0.36\% & -0.30\% & 0.26\% \\
    \hline
\end{tabular}}
\caption{Difference in counterfactual flip rates per technique on CivilComments compared to the original model.}
\label{table:counterfactual_flip_rates}
\end{table}

\subsection{Dataset Debiasing}
We replicate the experimental setting from \citep{dixon2018measuringmitigating} to evaluate dataset and model bias. We additionally augment their data augmentation techniques and introduce counterfactual generation to improve effectiveness of data debiasing and model remediation.

\subsubsection{Data debiasing techniques.} We use their keyword list as a baseline to understand toxicity rates, compute subgroup rate deficits and source non-toxic examples from Wikipedia article snippets for debiasing. We implement two additional techniques: i) sourcing using subgroup annotations instead of keywords and ii) generating five least similar counterfactual examples per label. We train a model per augmented dataset and evaluate classification performance on a templated synthetic dataset. Additional details can be found in Appendix \ref{appendix_data_models}.

\subsubsection{Dataset toxicity rates and model AUC.} 
\begin{figure}[htbp]
\centering
\includegraphics[width=0.9\columnwidth]{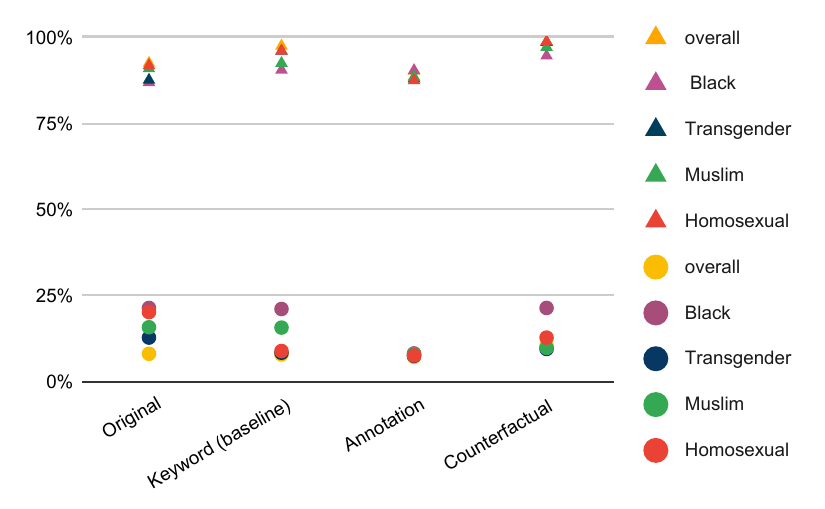} % Reduce the figure size so that it is slightly narrower than the column. Don't use precise values for figure width.This setup will avoid overfull boxes.
\caption{Model AUCs (triangles) and dataset toxicity rates (circles) per debiasing technique on a synthetic dataset. A tighter cluster pattern indicates less bias across subgroups.}
\label{fig:tox_rates_and_aucs}
\end{figure}

Annotation-driven data sourcing increases the coverage of terms compared to the keyword list, leading to more balanced toxicity rates across subgroups. Counterfactual augmentation increases per-label term diversity, resulting in the highest AUC scores and the most equality across subgroups in Figure~\ref{fig:tox_rates_and_aucs}. Toxicity rate balance from annotations translates to equality in model performance across subgroups, but with lower overall performance. 

\subsection{Generative Model Bias}
We replicate the experimental setting from \citep{smith-etal-2022-im} to evaluate generative model bias, leveraging our lexicon to expand the coverage of bias detection.

\subsubsection{Dataset generation.} We create two datasets: i) a baseline dataset using the templates and lexicon from HolisticBias and ii) a new dataset using our lexicon with the same templates. We generate perplexity scores by running evaluations of the 90M-paremeter BlenderBot model on both datasets.

\subsubsection{Token likelihood bias.} Our lexicon's deeper coverage of terms reveals a broader bias in token likelihoods for RNE in Figure~\ref{fig:generative_bias}. SOGIESC and Religion have a much smaller vocabulary as seen in Appendix \ref{appendix_tidal_distribution}, thus are not as prone to coverage issues.

\begin{figure}[htbp]
\centering
\begin{subfigure}[h]{0.45\linewidth}
\includegraphics[width=\linewidth]{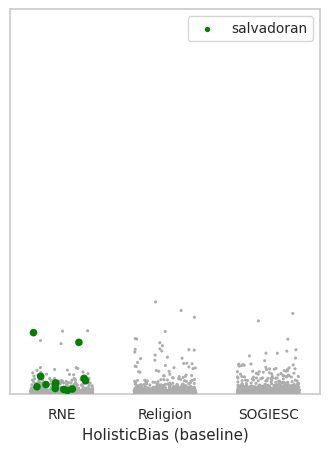} 
\end{subfigure}
\begin{subfigure}[h]{0.45\linewidth}
\includegraphics[width=\linewidth]{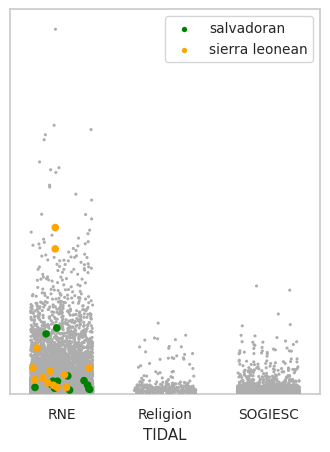} 
\end{subfigure}
\caption{Generative model perplexities on a synthetic dataset, with a max of 6000. Our lexicon shows an example of a previously missed term.}
\label{fig:generative_bias}
\end{figure}

\section{Conclusion}
We create a new identity lexicon, TIDAL and use it to develop an annotation tool for textual identity detection and augmentation. Through our experiments we demonstrate the effectiveness of our work to scale and improve existing human annotation and fairness techniques. 

When coupled with a comprehensive lexicon that includes term forms and expansions, token-matching emerges as the most practical annotation technique given its implementation simplicity and low computational cost. We note that a custom NER model results in computational speed gains, but requires training resources and ground truth annotations. We demonstrate improvements in human annotation reliability and cost, positioning our annotator as an assistive tool for acquiring identity labels from contributors. 

To scale fairness in practice, we build on our work to advance techniques used throughout the machine learning workflow. We demonstrate how to increase reliability in human annotations of ground truth, uncover more bias in data than previously known and train more fair models using improved techniques. We find that our approaches can be leveraged across different notions of fairness, ML development stages and model types. 

\section{Limitations}
Our current lexicon is limited in a number of ways due to the scope of the paper. We propose future work to increase the number of represented identity groups and subgroups. The scope of terms can be expanded to include non-literal associative words (e.g. ``temple'' for Religion), compound phrases that imply an identity group (e.g. ``same-sex marriage'' for SOGIEC), and prevalent stereotypes (e.g. ``kinky hair'' for RNE), all the while considering intersectionality.  Coverage of contextual dimensions (Appendix~\ref{appendix_tidal_distribution}) can be improved for balance across groups. Additional sense context can also be added to improve disambiguation, for example by integrating with other lexical-semantic datasets such as WordNet and Wiktionary \citep{eckle-kohler-etal-2012-uby} as shown in Appendix~\ref{appendix_tidal_schema_design}

Token-based techniques presented are limited due to complexity of identity, contextual interpretation and fluidity of language. In addition to NLP, advanced knowledge-based approaches \citep{agirre-etal-2014-random} need to be explored for disambiguated identity detection. Generative techniques like DataSynth\footnote{\url{https://github.com/Tobiadefami/datasynth}} hold a lot of promise for counterfactual generation.  All of these require expanding the lexicon to include more ``sense context'' as mentioned above.

Our results show that trade-offs are required in fairness depending on use case and type of bias, as techniques have different impacts in datasets and models \citep{goldfarb-tarrant-etal-2021-intrinsic}. While our experiments use techniques independently, we propose future work to examine mixed-method approaches to improve guidelines for practical settings.

Finally, our goal is to incorporate sense context from many perspectives, however crowd-sourcing does not explicitly advance this goal. Contributor diversity, task sensitivity and a lack of benchmarks all impact representation and perceived quality. Future work on identity datasets should explore participatory data collection and governance models to empower groups to not only shape how they're represented, but also where and how their data is used.

\section{Ethical Statement}
During our research we encounter a variety of questions, including how to collect identity context data ethically, how assistive context could bias human annotations, and what the right compensation for those tasks should be. 

We acknowledge that there are a lot more demographic categories and context than we choose to focus on in this paper. This means the work presented does not mitigate bias for everyone. Given our limited scope there is a high risk of misrepresentation and disenfranchisement especially of historically underrepresented groups.

We recommend caution when generalizing our findings to non-English languages or even across different cultures and groups given the subjectivity of identity assertions and toxicity labels. 

\subsection{Wellness in Human Evaluation}
Toxicity labeling has a side-effect of exposing human annotators and researchers to toxic languages, something we experience first-hand during our work. We only select contributors that accept explicit content \citep{GuideToC47:online} on the Appen platform. 

We also leverage the Fair Pay plugin \citep{GuidetoF30:online} to ensure that each contributor is fairly compensated based on their geographical location, with an extra 50\% pay increase over the suggested baseline to account for task complexity.

\bibliography{anthology,custom}

\begin{thebibliography}{52}
\expandafter\ifx\csname natexlab\endcsname\relax\def\natexlab#1{#1}\fi

\bibitem[{Abadi et~al.(2015)Abadi, Agarwal, Barham, Brevdo, Chen, Citro,
  Corrado, Davis, Dean, Devin, Ghemawat, Goodfellow, Harp, Irving, Isard, Jia,
  Jozefowicz, Kaiser, Kudlur, Levenberg, Man\'{e}, Monga, Moore, Murray, Olah,
  Schuster, Shlens, Steiner, Sutskever, Talwar, Tucker, Vanhoucke, Vasudevan,
  Vi\'{e}gas, Vinyals, Warden, Wattenberg, Wicke, Yu, and
  Zheng}]{tensorflow2015-whitepaper}
Mart\'{i}n Abadi, Ashish Agarwal, Paul Barham, Eugene Brevdo, Zhifeng Chen,
  Craig Citro, Greg~S. Corrado, Andy Davis, Jeffrey Dean, Matthieu Devin,
  Sanjay Ghemawat, Ian Goodfellow, Andrew Harp, Geoffrey Irving, Michael Isard,
  Yangqing Jia, Rafal Jozefowicz, Lukasz Kaiser, Manjunath Kudlur, Josh
  Levenberg, Dandelion Man\'{e}, Rajat Monga, Sherry Moore, Derek Murray, Chris
  Olah, Mike Schuster, Jonathon Shlens, Benoit Steiner, Ilya Sutskever, Kunal
  Talwar, Paul Tucker, Vincent Vanhoucke, Vijay Vasudevan, Fernanda Vi\'{e}gas,
  Oriol Vinyals, Pete Warden, Martin Wattenberg, Martin Wicke, Yuan Yu, and
  Xiaoqiang Zheng. 2015.
\newblock \href {https://www.tensorflow.org/} {{TensorFlow}: Large-scale
  machine learning on heterogeneous systems}.
\newblock Software available from tensorflow.org.

\bibitem[{Abid et~al.(2021)Abid, Farooqi, and Zou}]{abid2021persistent}
Abubakar Abid, Maheen Farooqi, and James Zou. 2021.
\newblock \href {https://doi.org/10.1145/3461702.3462624} {Persistent
  anti-muslim bias in large language models}.
\newblock In \emph{Proceedings of the 2021 AAAI/ACM Conference on AI, Ethics,
  and Society}, pages 298--306.

\bibitem[{Agirre et~al.(2014)Agirre, de~Lacalle, and
  Soroa}]{agirre-etal-2014-random}
Eneko Agirre, Oier~L{\'o}pez de~Lacalle, and Aitor Soroa. 2014.
\newblock \href {https://doi.org/10.1162/COLI_a_00164} {Random walks for
  knowledge-based word sense disambiguation}.
\newblock \emph{Computational Linguistics}, 40(1):57--84.

\bibitem[{Allan(2007)}]{pragmatics2007}
Keith Allan. 2007.
\newblock \href {https://doi.org/10.1016/j.pragma.2006.08.004} {The pragmatics
  of connotation}.
\newblock \emph{Journal of Pragmatics}, 39:1047--1057.

\bibitem[{Allaway and McKeown(2021)}]{allaway-mckeown-2021-unified}
Emily Allaway and Kathleen McKeown. 2021.
\newblock \href {https://doi.org/10.18653/v1/2021.eacl-main.184} {A unified
  feature representation for lexical connotations}.
\newblock In \emph{Proceedings of the 16th Conference of the European Chapter
  of the Association for Computational Linguistics: Main Volume}, pages
  2145--2163, Online. Association for Computational Linguistics.

\bibitem[{Andrus et~al.(2021)Andrus, Spitzer, Brown, and
  Xiang}]{andrus-spitzer-2021}
McKane Andrus, Elena Spitzer, Jeffrey Brown, and Alice Xiang. 2021.
\newblock \href {https://doi.org/10.1145/3442188.3445888} {What we can't
  measure, we can't understand: Challenges to demographic data procurement in
  the pursuit of fairness}.
\newblock In \emph{Proceedings of the 2021 ACM Conference on Fairness,
  Accountability, and Transparency}, page 249–260.

\bibitem[{Andrus and Villeneuve(2022)}]{andrus-villeneuve-2022}
McKane Andrus and Sarah Villeneuve. 2022.
\newblock \href {https://doi.org/10.1145/3531146.3533226} {Demographic-reliant
  algorithmic fairness: Characterizing the risks of demographic data collection
  in the pursuit of fairness}.
\newblock In \emph{Proceedings of the 2022 ACM Conference on Fairness,
  Accountability, and Transparency}, page 1709–1721.

\bibitem[{Appen({\natexlab{a}})}]{GuideToC47:online}
Appen. {\natexlab{a}}.
\newblock \href
  {https://success.appen.com/hc/en-us/articles/203219195-Guide-To-Contributors-Channels-Page}
  {Guide to: Contributors channels page}.
\newblock [Online; accessed 28-June-2023].

\bibitem[{Appen({\natexlab{b}})}]{GuideToD99:online}
Appen. {\natexlab{b}}.
\newblock \href
  {https://success.appen.com/hc/en-us/articles/13741727264909-Guide-To-Designating-Test-Questions-for-Quiz-or-Work-Mode-with-Gold-Pool}
  {Guide to: Designating test questions for quiz or work mode with gold pool}.
\newblock [Online; accessed 28-June-2023].

\bibitem[{Appen({\natexlab{c}})}]{GuidetoF30:online}
Appen. {\natexlab{c}}.
\newblock \href
  {https://success.appen.com/hc/en-us/articles/9557008940941-Guide-to-Fair-Pay}
  {Guide to: Fair pay}.
\newblock [Online; accessed 28-June-2023].

\bibitem[{Appen({\natexlab{d}})}]{MachineL45:online}
Appen. {\natexlab{d}}.
\newblock \href
  {https://success.appen.com/hc/en-us/articles/360030714912-Machine-Learning-Assisted-Text-Utterance-Collection-Model-Details}
  {Machine learning assisted text utterance collection}.
\newblock [Online; accessed 28-June-2023].

\bibitem[{Bellamy et~al.(2018)Bellamy, Dey, Hind, Hoffman, Houde, Kannan,
  Lohia, Martino, Mehta, Mojsilovic, Nagar, Ramamurthy, Richards, Saha,
  Sattigeri, Singh, Varshney, and Zhang}]{bellamy2018ai}
Rachel K.~E. Bellamy, Kuntal Dey, Michael Hind, Samuel~C. Hoffman, Stephanie
  Houde, Kalapriya Kannan, Pranay Lohia, Jacquelyn Martino, Sameep Mehta,
  Aleksandra Mojsilovic, Seema Nagar, Karthikeyan~Natesan Ramamurthy, John~T.
  Richards, Diptikalyan Saha, Prasanna Sattigeri, Moninder Singh, Kush~R.
  Varshney, and Yunfeng Zhang. 2018.
\newblock \href {http://arxiv.org/abs/1810.01943} {{AI} fairness 360: An
  extensible toolkit for detecting, understanding, and mitigating unwanted
  algorithmic bias}.
\newblock \emph{Computing Research Repository}, abs/1810.01943.

\bibitem[{Bird et~al.(2020)Bird, Dud{\'i}k, Edgar, Horn, Lutz, Milan, Sameki,
  Wallach, and Walker}]{bird2020fairlearn}
Sarah Bird, Miro Dud{\'i}k, Richard Edgar, Brandon Horn, Roman Lutz, Vanessa
  Milan, Mehrnoosh Sameki, Hanna Wallach, and Kathleen Walker. 2020.
\newblock \href
  {https://www.microsoft.com/en-us/research/publication/fairlearn-a-toolkit-for-assessing-and-improving-fairness-in-ai/}
  {Fairlearn: A toolkit for assessing and improving fairness in {AI}}.
\newblock Technical Report MSR-TR-2020-32, Microsoft.

\bibitem[{Bird et~al.(2009)Bird, Klein, and Loper}]{nltkbook}
Steven Bird, Ewan Klein, and Edward Loper. 2009.
\newblock \emph{Natural language processing with Python: analyzing text with
  the natural language toolkit}.
\newblock O'Reilly Media, Inc.

\bibitem[{Blodgett et~al.(2020)Blodgett, Barocas, Daum{\'e}~III, and
  Wallach}]{blodgett-etal-2020-language}
Su~Lin Blodgett, Solon Barocas, Hal Daum{\'e}~III, and Hanna Wallach. 2020.
\newblock \href {https://doi.org/10.18653/v1/2020.acl-main.485} {Language
  (technology) is power: A critical survey of {``}bias{''} in {NLP}}.
\newblock In \emph{Proceedings of the 58th Annual Meeting of the Association
  for Computational Linguistics}, pages 5454--5476, Online. Association for
  Computational Linguistics.

\bibitem[{Borkan et~al.(2019)Borkan, Dixon, Sorensen, Thain, and
  Vasserman}]{borkan2019nuanced}
Daniel Borkan, Lucas Dixon, Jeffrey Sorensen, Nithum Thain, and Lucy Vasserman.
  2019.
\newblock \href {https://doi.org/10.1145/3308560.3317593} {Nuanced metrics for
  measuring unintended bias with real data for text classification}.
\newblock In \emph{Companion Proceedings of The 2019 World Wide Web
  Conference}, pages 491--500.

\bibitem[{Carpuat(2015)}]{carpuat-2015-connotation}
Marine Carpuat. 2015.
\newblock \href {https://doi.org/10.18653/v1/W15-2903} {Connotation in
  translation}.
\newblock In \emph{Proceedings of the 6th Workshop on Computational Approaches
  to Subjectivity, Sentiment and Social Media Analysis}, pages 9--15, Lisboa,
  Portugal. Association for Computational Linguistics.

\bibitem[{Celis et~al.(2021)Celis, Huang, Keswani, and Vishnoi}]{celis2021fair}
L~Elisa Celis, Lingxiao Huang, Vijay Keswani, and Nisheeth~K Vishnoi. 2021.
\newblock \href {https://proceedings.mlr.press/v139/celis21a.html} {Fair
  classification with noisy protected attributes: A framework with provable
  guarantees}.
\newblock In \emph{Proceedings of the 38th International Conference on Machine
  Learning}, volume 139, pages 1349--1361. PMLR.

\bibitem[{Dixon et~al.(2018)Dixon, Li, Sorensen, Thain, and
  Vasserman}]{dixon2018measuringmitigating}
Lucas Dixon, John Li, Jeffrey Sorensen, Nithum Thain, and Vasserman. 2018.
\newblock \href {https://doi.org/10.1145/3278721.3278729} {Measuring and
  mitigating unintended bias in text classification}.
\newblock In \emph{Proceedings of the 2018 AAAI/ACM Conference on AI, Ethics,
  and Society}, AIES '18, page 67–73.

\bibitem[{Eckle-Kohler et~al.(2012)Eckle-Kohler, Gurevych, Hartmann, Matuschek,
  and Meyer}]{eckle-kohler-etal-2012-uby}
Judith Eckle-Kohler, Iryna Gurevych, Silvana Hartmann, Michael Matuschek, and
  Christian~M. Meyer. 2012.
\newblock \href
  {http://www.lrec-conf.org/proceedings/lrec2012/pdf/475_Paper.pdf}
  {{UBY}-{LMF} {--} a uniform model for standardizing heterogeneous
  lexical-semantic resources in {ISO}-{LMF}}.
\newblock In \emph{Proceedings of the Eighth International Conference on
  Language Resources and Evaluation ({LREC}'12)}, pages 275--282, Istanbul,
  Turkey. European Language Resources Association (ELRA).

\bibitem[{for Standardization(2022)}]{ISO2461372:online}
International~Organization for Standardization. 2022.
\newblock \href {https://www.iso.org/standard/72099.html} {Language resource
  management — lexical markup framework (lmf) — part 5: Lexical base
  exchange (lbx) serialization}.
\newblock ISO Standard No. 24613-5:2022.

\bibitem[{Garg et~al.(2019)Garg, Perot, Limtiaco, Taly, Chi, and
  Beutel}]{garg2019counterfactual}
Sahaj Garg, Vincent Perot, Nicole Limtiaco, Ankur Taly, Ed~H Chi, and Alex
  Beutel. 2019.
\newblock \href {https://doi.org/10.1145/3306618.3317950} {Counterfactual
  fairness in text classification through robustness}.
\newblock In \emph{Proceedings of the 2019 AAAI/ACM Conference on AI, Ethics,
  and Society}, pages 219--226.

\bibitem[{Gerner et~al.(2002)Gerner, Abu-Jabr, Schrodt, and
  Yilmaz}]{Gerner2002ConflictAM}
Deborah~J. Gerner, Rajaa Abu-Jabr, Philip~A. Schrodt, and {\"O}m{\"u}r Yilmaz.
  2002.
\newblock \href {https://parusanalytics.com/eventdata/papers.dir/gerner02.pdf}
  {Conflict and mediation event observations (cameo): A new event data
  framework for the analysis of foreign policy interactions}.
\newblock \emph{International Studies Association}.
\newblock Paper presented at the 43rd Annual Convention of the International
  Studies Association, New Orleans, March 2002.

\bibitem[{{GLAAD}()}]{GLAADGlossary:online}
{GLAAD}.
\newblock \href {https://glaad.org/reference/terms/} {Glossary of terms: Lgbtq
  - glaad}.
\newblock [Online; accessed 28-June-2023].

\bibitem[{Goldfarb-Tarrant et~al.(2021)Goldfarb-Tarrant, Marchant,
  Mu{\~n}oz~S{\'a}nchez, Pandya, and
  Lopez}]{goldfarb-tarrant-etal-2021-intrinsic}
Seraphina Goldfarb-Tarrant, Rebecca Marchant, Ricardo Mu{\~n}oz~S{\'a}nchez,
  Mugdha Pandya, and Adam Lopez. 2021.
\newblock \href {https://doi.org/10.18653/v1/2021.acl-long.150} {Intrinsic bias
  metrics do not correlate with application bias}.
\newblock In \emph{Proceedings of the 59th Annual Meeting of the Association
  for Computational Linguistics and the 11th International Joint Conference on
  Natural Language Processing (Volume 1: Long Papers)}, pages 1926--1940,
  Online. Association for Computational Linguistics.

\bibitem[{Gupta et~al.(2018)Gupta, Cotter, Fard, and Wang}]{gupta2018proxy}
Maya~R. Gupta, Andrew Cotter, Mahdi~Milani Fard, and Serena~Lutong Wang. 2018.
\newblock \href {http://arxiv.org/abs/1806.11212} {Proxy fairness}.
\newblock \emph{Computing Research Repository}, abs/1806.11212.

\bibitem[{Gwet(2014)}]{gwet2014handbook}
Kilem~L Gwet. 2014.
\newblock \emph{Handbook of inter-rater reliability: The definitive guide to
  measuring the extent of agreement among raters}.
\newblock Advanced Analytics, LLC.

\bibitem[{Honnibal and Montani(2017)}]{spacy2}
Matthew Honnibal and Ines Montani. 2017.
\newblock \href {https://spacy.io} {{spaCy 2}: Natural language understanding
  with {B}loom embeddings, convolutional neural networks and incremental
  parsing}.
\newblock To appear.

\bibitem[{{HRC Foundation}()}]{HRCGlossary:online}
{HRC Foundation}.
\newblock \href {https://www.hrc.org/resources/glossary-of-terms} {Glossary of
  terms - human rights campaign}.
\newblock [Online; accessed 28-June-2023].

\bibitem[{Jigsaw(2019)}]{JigsawUn23:online}
Jigsaw. 2019.
\newblock \href
  {https://www.kaggle.com/competitions/jigsaw-unintended-bias-in-toxicity-classification/data}
  {Jigsaw unintended bias in toxicity classification}.
\newblock [Online; accessed 28-June-2023].

\bibitem[{Jung et~al.(2022)Jung, Chun, and Moon}]{jung2022learning}
Sangwon Jung, Sanghyuk Chun, and Taesup Moon. 2022.
\newblock \href {https://doi.org/10.1109/CVPR52688.2022.01010} {Learning fair
  classifiers with partially annotated group labels}.
\newblock In \emph{Proceedings of the IEEE/CVF Conference on Computer Vision
  and Pattern Recognition}, pages 10348--10357.

\bibitem[{Krippendorff(1970)}]{krippendorff1970estimating}
Klaus Krippendorff. 1970.
\newblock \href {https://doi.org/10.1177/001316447003000105} {Estimating the
  reliability, systematic error and random error of interval data}.
\newblock \emph{Educational and psychological measurement}, 30(1):61--70.

\bibitem[{Lacy et~al.(2015)Lacy, Watson, Riffe, and Lovejoy}]{lacy2015issues}
Stephen Lacy, Brendan~R. Watson, Daniel Riffe, and Jennette Lovejoy. 2015.
\newblock \href {https://doi.org/10.1177/1077699015607338} {Issues and best
  practices in content analysis}.
\newblock \emph{Journalism \& Mass Communication Quarterly}, 92(4):791--811.

\bibitem[{Lahoti et~al.(2020)Lahoti, Beutel, Chen, Lee, Prost, Thain, Wang, and
  Chi}]{lahoti2020fairness}
Preethi Lahoti, Alex Beutel, Jilin Chen, Kang Lee, Flavien Prost, Nithum Thain,
  Xuezhi Wang, and Ed~Chi. 2020.
\newblock \href
  {https://proceedings.neurips.cc/paper/2020/file/07fc15c9d169ee48573edd749d25945d-Paper.pdf}
  {Fairness without demographics through adversarially reweighted learning}.
\newblock In \emph{Proceedings of the 34th International Conference on Neural
  Information Processing Systems}, pages 728--740.

\bibitem[{McLoughney et~al.(2023)McLoughney, Paterson, Cheong, and
  Wirth}]{mcloughney2023}
Aidan~James McLoughney, Jeannie~Marie Paterson, Marc Cheong, and Anthony Wirth.
  2023.
\newblock \href {https://doi.org/10.1109/ETHICS57328.2023.10155045}
  {‘emerging proxies’ in information-rich machine learning: a threat to
  fairness?}
\newblock In \emph{2023 IEEE International Symposium on Ethics in Engineering,
  Science, and Technology (ETHICS)}, pages 1--1.

\bibitem[{Morning(2008)}]{morning2008ethnic}
Ann Morning. 2008.
\newblock \href {https://doi.org/10.1007/s11113-007-9062-5} {Ethnic
  classification in global perspective: A cross-national survey of the 2000
  census round}.
\newblock \emph{Population Research and Policy Review}, 27:239--272.

\bibitem[{Pal and Saha(2015)}]{pal2015word}
Alok~Ranjan Pal and Diganta Saha. 2015.
\newblock \href {https://doi.org/10.5121/ijctcm.2015.5301} {Word sense
  disambiguation: A survey}.
\newblock \emph{International Journal of Control Theory and Computer Modeling
  (IJCTCM)}, 5(3).

\bibitem[{Raffel et~al.(2020)Raffel, Shazeer, Roberts, Lee, Narang, Matena,
  Zhou, Li, and Liu}]{c4:2020}
Colin Raffel, Noam Shazeer, Adam Roberts, Katherine Lee, Sharan Narang, Michael
  Matena, Yanqi Zhou, Wei Li, and Peter~J. Liu. 2020.
\newblock \href {http://jmlr.org/papers/v21/20-074.html} {Exploring the limits
  of transfer learning with a unified text-to-text transformer}.
\newblock \emph{Journal of Machine Learning Research}, 21(140):1--67.

\bibitem[{Roller et~al.(2021)Roller, Dinan, Goyal, Ju, Williamson, Liu, Xu,
  Ott, Smith, Boureau, and Weston}]{roller-etal-2021-recipes}
Stephen Roller, Emily Dinan, Naman Goyal, Da~Ju, Mary Williamson, Yinhan Liu,
  Jing Xu, Myle Ott, Eric~Michael Smith, Y-Lan Boureau, and Jason Weston. 2021.
\newblock \href {https://doi.org/10.18653/v1/2021.eacl-main.24} {Recipes for
  building an open-domain chatbot}.
\newblock In \emph{Proceedings of the 16th Conference of the European Chapter
  of the Association for Computational Linguistics: Main Volume}, pages
  300--325, Online. Association for Computational Linguistics.

\bibitem[{Ross et~al.(2016)Ross, Rist, Carbonell, Cabrera, Kurowsky, and
  Wojatzki}]{ross2017measuring}
Bj{\"o}rn Ross, Michael Rist, Guillermo Carbonell, Benjamin Cabrera, Nils
  Kurowsky, and Michael Wojatzki. 2016.
\newblock \href
  {https://www.research.ed.ac.uk/en/publications/measuring-the-reliability-of-hate-speech-annotations-the-case-of-}
  {Measuring the reliability of hate speech annotations: The case of the
  european refugee crisis}.
\newblock In \emph{NLP4CMC III: 3rd Workshop on Natural Language Processing for
  Computer-Mediated Communication}, pages 6--9. Ruhr-Universitat Bochum.

\bibitem[{Saleiro et~al.(2018)Saleiro, Kuester, Hinkson, London, Stevens,
  Anisfeld, Rodolfa, and Ghani}]{saleiro2018aequitas}
Pedro Saleiro, Benedict Kuester, Loren Hinkson, Jesse London, Abby Stevens, Ari
  Anisfeld, Kit~T Rodolfa, and Rayid Ghani. 2018.
\newblock \href {http://arxiv.org/abs/1811.05577} {Aequitas: A bias and
  fairness audit toolkit}.
\newblock \emph{Computing Research Repository}, abs/1811.05577.

\bibitem[{Shah et~al.(2020)Shah, Schwartz, and
  Hovy}]{shah-etal-2020-predictive}
Deven~Santosh Shah, H.~Andrew Schwartz, and Dirk Hovy. 2020.
\newblock \href {https://doi.org/10.18653/v1/2020.acl-main.468} {Predictive
  biases in natural language processing models: A conceptual framework and
  overview}.
\newblock In \emph{Proceedings of the 58th Annual Meeting of the Association
  for Computational Linguistics}, pages 5248--5264, Online. Association for
  Computational Linguistics.

\bibitem[{Smith et~al.(2022)Smith, Hall, Kambadur, Presani, and
  Williams}]{smith-etal-2022-im}
Eric~Michael Smith, Melissa Hall, Melanie Kambadur, Eleonora Presani, and Adina
  Williams. 2022.
\newblock \href {https://aclanthology.org/2022.emnlp-main.625} {{``}{I}{'}m
  sorry to hear that{''}: Finding new biases in language models with a holistic
  descriptor dataset}.
\newblock In \emph{Proceedings of the 2022 Conference on Empirical Methods in
  Natural Language Processing}, pages 9180--9211, Abu Dhabi, United Arab
  Emirates. Association for Computational Linguistics.

\bibitem[{Sweeney(2013)}]{sweeney2013discrimination}
Latanya Sweeney. 2013.
\newblock \href {https://doi.org/10.1145/2447976.2447990} {Discrimination in
  online ad delivery}.
\newblock \emph{Communications of the ACM}, 56(5):44--54.

\bibitem[{Trithart(2021)}]{trithart2021}
Albert Trithart. 2021.
\newblock \href {https://www.jstor.org/stable/resrep28857} {A un for all?: Un
  policy and programming on sexual orientation, gender identity and expression,
  and sex characteristics}.
\newblock Technical report, International Peace Institute.

\bibitem[{Tschantz(2022)}]{tschantz2022}
Michael~Carl Tschantz. 2022.
\newblock \href {https://doi.org/10.1145/3531146.3533242} {What is proxy
  discrimination?}
\newblock In \emph{2022 ACM Conference on Fairness, Accountability, and
  Transparency}, FAccT '22, page 1993–2003, New York, NY, USA. Association
  for Computing Machinery.

\bibitem[{UNSD(2003)}]{united2003ethnicity}
UNSD. 2003.
\newblock \href
  {https://unstats.un.org/unsd/demographic/sconcerns/popchar/ethnicitypaper.pdf}
  {Ethnicity: A review of data collection and dissemination}.
\newblock Technical report, United Nations Statistics Division.

\bibitem[{Wadhwa et~al.(2022)Wadhwa, Bhambhani, Jindal, Sawant, and
  Madhavan}]{wadhwa2022fairness}
Mohit Wadhwa, Mohan Bhambhani, Ashvini Jindal, Uma Sawant, and Ramanujam
  Madhavan. 2022.
\newblock \href {http://arxiv.org/abs/2203.03541} {Fairness for text
  classification tasks with identity information data augmentation methods}.
\newblock \emph{Computing Research Repository}, abs/2203.03541.
\newblock Paper presented at the Measures and Best Practices for Responsible AI
  workshop at the 27th ACM SIGKDD Conference on Knowledge Discovery \& Data
  Mining, Virtual/Singapore, August 2021.

\bibitem[{Webson et~al.(2020)Webson, Chen, Eickhoff, and
  Pavlick}]{webson-etal-2020-undocumented}
Albert Webson, Zhizhong Chen, Carsten Eickhoff, and Ellie Pavlick. 2020.
\newblock \href {https://doi.org/10.18653/v1/2020.emnlp-main.335} {Are
  {``}undocumented workers{''} the same as {``}illegal aliens{''}?
  {D}isentangling denotation and connotation in vector spaces}.
\newblock In \emph{Proceedings of the 2020 Conference on Empirical Methods in
  Natural Language Processing (EMNLP)}, pages 4090--4105, Online. Association
  for Computational Linguistics.

\bibitem[{{Wikipedia contributors}(2023)}]{WikipediaDemonyms:online}
{Wikipedia contributors}. 2023.
\newblock \href
  {https://en.wikipedia.org/w/index.php?title=List_of_adjectival_and_demonymic_forms_for_countries_and_nations&oldid=1157246055}
  {List of adjectival and demonymic forms for countries and nations ---
  {Wikipedia}{,} the free encyclopedia}.
\newblock [Online; accessed 28-June-2023].

\bibitem[{{Wiktionary contributors}(2021)}]{WikiTermsOfAddress:online}
{Wiktionary contributors}. 2021.
\newblock \href
  {https://en.wiktionary.org/w/index.php?title=Category:English_terms_of_address&oldid=63763769}
  {Category - english terms of address --- {Wiktionary}{,} the free
  dictionary}.
\newblock [Online; accessed 28-June-2023].

\bibitem[{Wulczyn et~al.(2017)Wulczyn, Thain, and Dixon}]{wulczyn2017ex}
Ellery Wulczyn, Nithum Thain, and Lucas Dixon. 2017.
\newblock \href {https://doi.org/10.1145/3038912.3052591} {Ex machina: Personal
  attacks seen at scale}.
\newblock In \emph{Proceedings of the 26th international conference on world
  wide web}, pages 1391--1399.

\end{thebibliography}
\bibliographystyle{acl_natbib}

\begin{appendices}
\setcounter{secnumdepth}{2} %May be changed to 1 or 2 if section numbers are desired.

\appendixpage
\textcolor{red}{\textit{Warning: Following sections may include terms or generated sample text that may be considered offensive or toxic.}}
\section{TIDAL Dataset}
\subsection{Schema Design} \label{appendix_tidal_schema_design}
\begin{figure}[htbp]
\centering
\includegraphics[width=\columnwidth]{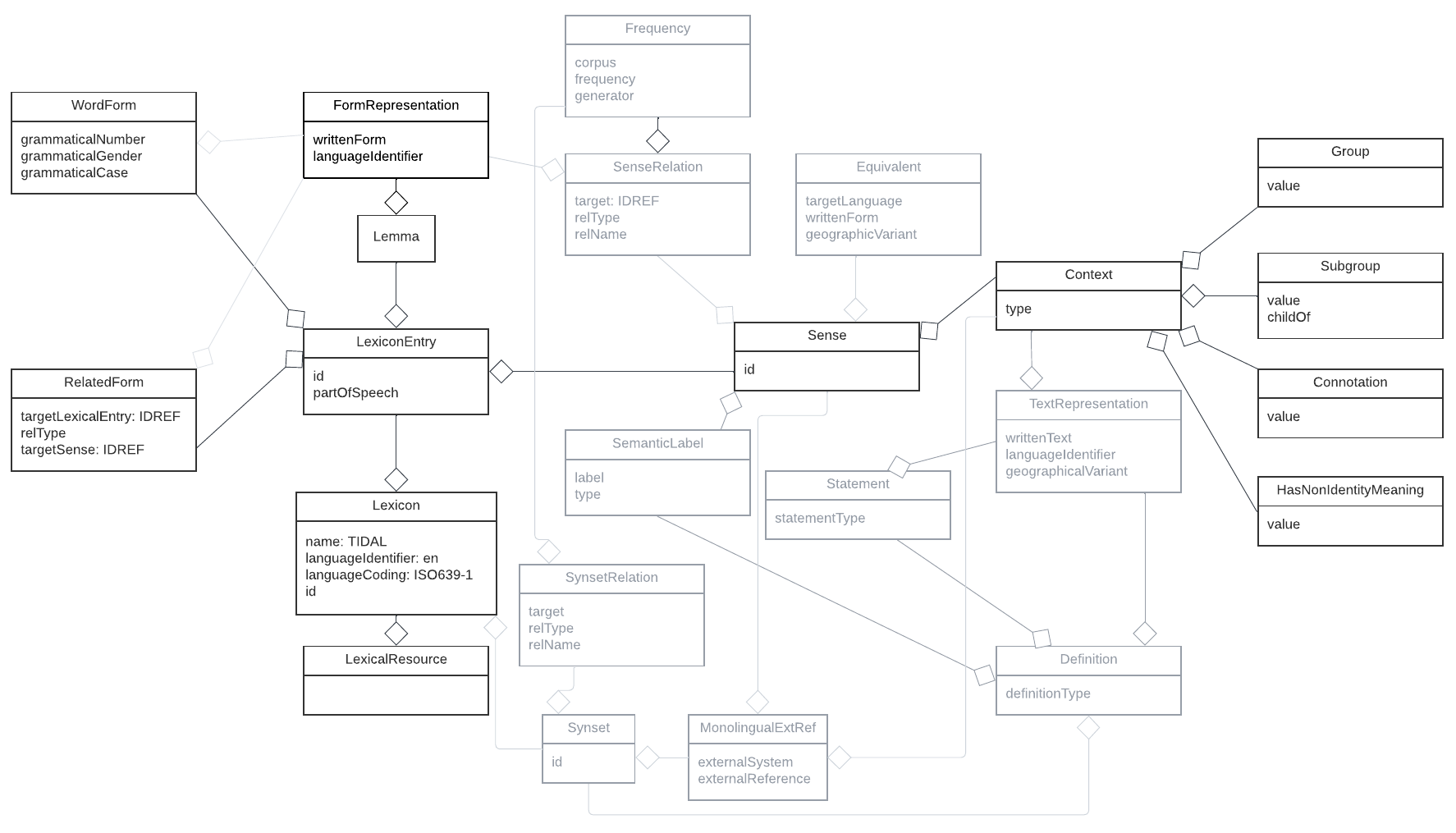}
\caption{Simplified Entity-relationship diagram of the lexicon schema using UML notation.}
\label{fig:lexicon_uml_diagram}
\end{figure}

We leverage and customize the UBY-LMF model \citep{eckle-kohler-etal-2012-uby} for its comprehensiveness and extensibility in supporting lexical, semantic and pragmatic properties of words and phrases. Figure~\ref{fig:lexicon_uml_diagram} shows a simplified Entity-relationship diagram (ERD) of the lexicon schema using UML notation. The grayed out entities and relationship are not in scope of this paper, but are shown in the diagram to support the extensibility argument for choosing the UBY-LMF model for our schema. 

This schema allows us to model lexical information types in detail, including morphology, syntax, semantic and pragmatic arguments. It also enables standard-compliant sense alignments between other lexical sources. We define subclasses for the Context class, allowing us to model the context as a subclass of Sense entry associated with the lexical entry.

\subsection{Post-processing of seed set} \label{appendix_post_processing_seed_set}
All seed set sources are cleaned-up by lowercasing the data, removing punctuations, numbers, extra space, hyphens and back/forward slashes. Additionally for RNE category, we use country names from Wikipedia as a filter to remove terms which could be country names from the seed set. All sources are then aggregated and provenance of sourcing and post-processing is stored along with the seed term set. These seed terms then form the lexical entries for our lexicon.

\subsection{Data distribution of TIDAL} \label{appendix_tidal_distribution}
Table~\ref{table:tidal_distribution_1} shows the distribution of TIDAL across IdentityGroups while Table~\ref{table:tidal_distribution_2} shows the distribution across Connotation context.

\begin{table}[htbp]
\resizebox{.95\columnwidth}{!}{\begin{tabular}{|p{1in}|c|c|c|c|}\hline
     & \textbf{Total} & \textbf{RNE} & \textbf{Religion} & \textbf{SOGIESC} \\
    \hline
    All Entries & 15123 & 13762 & 355 & 1046 \\
    \hline
    Head Entries & 1277 & 1278 & 25 & 121 \\
    \hline
    Person Noun Compound Entries & 10090 & 9256 & 260 & 600 \\
    \hline
    Other Related Form Entries & 3592 & 3233 & 70 & 299 \\
    \hline
\end{tabular}}
\caption{TIDAL: Head Lexical entry and Related form distribution by IdentityGroup.}
\label{table:tidal_distribution_1}
\end{table}

\begin{table}[htbp]
\resizebox{.95\columnwidth}{!}{\begin{tabular}{|p{1in}|c|c|c|c|}\hline
     & \textbf{Total} & \textbf{RNE} & \textbf{Religion} & \textbf{SOGIESC} \\
    \hline
    All Entries & 15123 & 13762 & 355 & 1046 \\
    \hline
    NEUTRAL & 15031 & 13734 & 355 & 1054 \\
    \hline
    PEJORATIVE & 216 & 113 & 34 & 137 \\
    \hline
    BOTH & 124 & 30 & 17 & 60 \\
    \hline
\end{tabular}}
\caption{TIDAL: Connotation distribution by IdentityGroup.}
\label{table:tidal_distribution_2}
\end{table}

\section{Acquiring Identity Context} 
\subsection{Annotation tool design details} \label{appendix_annotation_details}
\subsubsection{Training and test data preprocessing.} We use the ``train'' split of the CivilComments dataset because other splits do not have identity annotations. We only include identity and toxicity labels where rater agreement is greater than 0.5.  We then partition the dataset using a 3-1 ratio for training (75\%) and test (25\%) data.  The test data partition is then used for evaluation of annotators. For C4, we use the ``validation'' split for evaluation of annotators.

\subsubsection{Custom NER model training.}
During qualitative analysis we observe some incorrect human-annotated labels on the CivilComment dataset. To ensure high-quality training data, we first annotate CivilComments using the exact-match annotator. We only use a label set as ground truth when the annotation tool matches human-annotated labels. We train a spaCy pipeline for 11 epochs with a 50\% dropout rate.

\subsection{Annotation tool results} \label{appendix_annotation_results}
\subsubsection{False Positives/False Negatives.} Analysis in Table ~\ref{table:annotation_fp_fn} shows a false positive, false negative tradeoff between token-matching and token-matching with disambiguation for RNE and SOGIESC. We however observe consistent false negatives for Religion across all annotators except exact-matching.

\begin{table}[htbp]
\resizebox{.95\columnwidth}{!}{\begin{tabular}{|p{1in}|cc|cc|cc|}\hline
     & \multicolumn{2}{|c|}{\textbf{RNE}} & \multicolumn{2}{|c|}{\textbf{Religion}} & \multicolumn{2}{|c|}{\textbf{SOGIESC}} \\
     & FP & FN & FP & FN & FP & FN \\
    \hline
    Substring match (baseline) & 24665 & 10 & 424 & 1572 & 15657 & 34 \\
    \hline
    Exact match on all term variants & 4523	& 27 & 206 & 804 & 2511 & 125 \\
    \hline
    Lemma match on head terms & 4079 & 32 & 197 & 1298 & 2270 & 103 \\
    \hline
    Lemma+lexicon-person filter & 2764 & 586 & 164 & 1571 & 2214 & 240 \\
    \hline
    Lemma+similarity-person filter & 3309 & 570 & 173 & 1126 & 2456 & 279 \\
    \hline
    Custom NER model & 3421 & 38 & 193 & 1217 & 2185 & 131 \\
    \hline
\end{tabular}}
\caption{Multi-class false positive (FP) and false negative (FN) counts for the annotation tool on CivilComments}
\label{table:annotation_fp_fn}
\end{table}

\subsubsection{C4 annotation tool performance as control}
\begin{figure}
\centering
\begin{subfigure}[t]{0.4\textwidth}
\includegraphics[width=\linewidth]{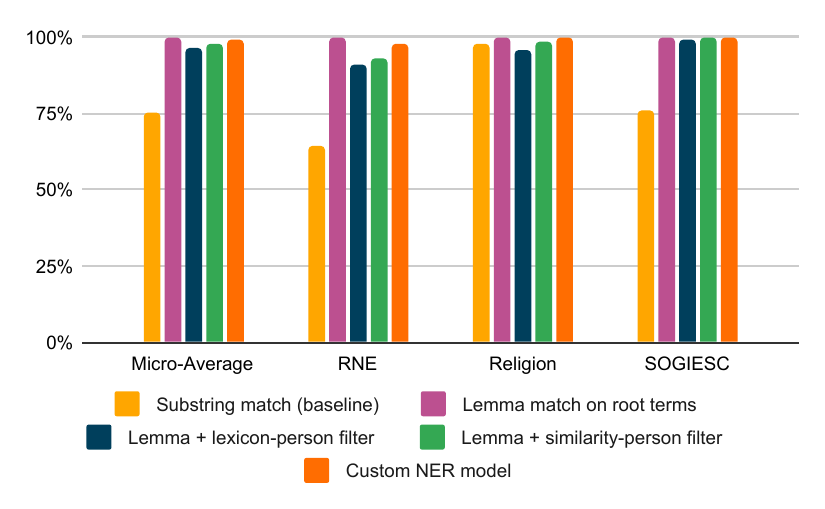} 
\end{subfigure}
\begin{subfigure}[t]{0.4\textwidth}
\includegraphics[width=\linewidth]{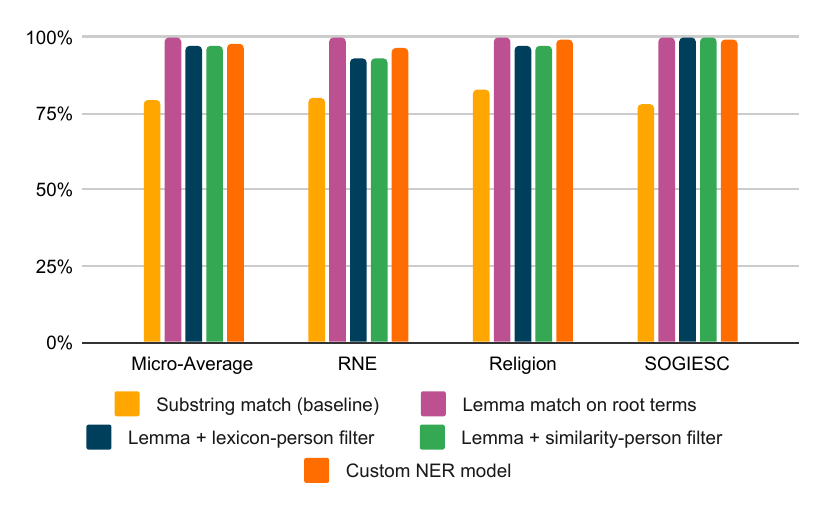} 
\end{subfigure}
\caption{Multi-class F1 scores for the annotation tool on CivilComments (top) and the C4 (bottom).}
\label{fig:appendix_annotator_performance}
\end{figure}

We corroborate annotation tool performance using a different dataset. We use C4 as the control, however it lacks human-annotated labels, so for consistency we treat the exact-match annotator as ground truth for both datasets in this evaluation. Results show similar performance on both datasets. Figure~\ref{fig:appendix_annotator_performance} shows overall and per-class performance for the annotators on both datasets. 

\subsection{Human computation design details} \label{appendix_hcomp_exp1_details}
The identity context human computation task is designed to be completed by crowd-sourced contributors on the data annotation platform. In the baseline task, contributors are asked to read the text and identify any tokens that they believe are associated with identity. In the assistive-annotation tasks, contributors are also provided with annotated associated class (RNE, Religion, or SOGIESC) for each identity token. The task consists of three steps:
\begin{enumerate}
    \item \textbf{Review the class labels.} Contributors are asked to review the class labels and definitions before beginning before beginning the actual annotation task. This helps ensure that contributors are familiar with the different types of classes and the criteria for annotation.
    \item \textbf{Read the text/comment}: We ask contributors to read the text/comment in detail.
    \item \textbf{Selecting/validating ``token'' and respective ``class''}: In the baseline task, contributors are asked to select any tokens that they believe are associated with identity. In the assistive-annotation task, they are asked to validate the assistive-annotations and select the ones that were missed.
\end{enumerate}

For each class of identity-related tokens, contributors are provided with specific guidelines and examples. For example, for the RNE class, they are asked to select tokens that refer to race, nationality or ethnicity (e.g. black, white, spaniard, indian) or RNE insults (e.g. wetback, bluegum). They are specifically instructed not to annotate people names (e.g. John, Abdul) or terms that do not describe a specific group's race or ethnicity (e.g. literal terms like racist, race, ethnicity, ethnic group).

Similarly, for the Religion class, contributors are asked to select tokens that refer to religious groups (e.g. islam, muslim, christian, jewish) or religious insults (e.g. kike, raghead). They are specifically instructed not to annotate people names/religious figures (e.g. Jesus, Christ, Mohammad, Bishop) or religious worship terms (e.g. Church, Temple, Mosque).

Finally for the SOGIESC class, contributors are asked to select tokens that refer to particular SOGIESC (e.g. trans, bisexual, cisgender, queer, lgbtq), SOGIESC insults (e.g. fag, poof, bull dyke) or gendered terms (e.g. man, woman). They are specifically instructed not to annotate pronouns (e.g. he, she, him, her, they), a gendered name (e.g. Donald, Margaret) or literal terms (e.g., sex, gender, sexual, sexist).

In addition to selecting and validating tokens, contributors are also asked to provide a brief explanation of why they believe the token is associated with the selected class. To help reduce spam and gibberish in this free-form text field, we use an ML-assisted text utterance tool by Appen on low threshold settings \citep{MachineL45:online}.

Finally, to ensure the quality of the annotations a small subset of the task questions are used in a test run. Questions with high agreement in their answers are then used as new test questions. We use the Gold pool feature by Appen \citep{GuideToD99:online} to select these test questions.

\subsection{Human computation results} \label{appendix_hcomp_exp1_results}
\subsubsection{IAR results.} Table~\ref{table:hcomp_exp1_iar_all_metrics} shows all measures we used for human annotation reliability evaluation.

\begin{table}[htbp]
\resizebox{.95\columnwidth}{!}{\begin{tabular}{|p{1in}|c|c|c|}\hline
     & \textbf{Percent Agreement} & \textbf{Krippendorff's Alpha} & \textbf{Gwet's AC1} \\
    \hline
    Example-only (baseline) & 0.4036 & 0.404 & 0.4027 \\
    \hline
    Assistive Identity Group annotations & 0.7636 & 0.763 & 0.7622 \\
    \hline
    Assistive Identity Group + Connotation annotations & 0.6265 & 0.6316 & 0.6257 \\
    \hline
\end{tabular}}
\caption{IAR for human annotations: identity labeling on CivilComments (All metrics)}
\label{table:hcomp_exp1_iar_all_metrics}
\end{table}

\subsubsection{False Positives/False Negatives.} Analysis in Table~\ref{table:exp1_fp_fn_cc} shows a false positive, false negative trade-off between assistive annotations with token-matching and with disambiguation for all 3 groups, while token-matching with disambiguation additionally had  No Classes false positives.

\begin{table}[htbp]
\resizebox{.95\columnwidth}{!}{\begin{tabular}{|p{1in}|cc|cc|cc|c|}\hline
     & \multicolumn{2}{|c|}{\textbf{RNE}} & \multicolumn{2}{|c|}{\textbf{Religion}} &
     \multicolumn{2}{|c|}{\textbf{SOGIESC}} &
     \multicolumn{1}{|c|}{\textbf{No Classes}} \\
     & FP & FN & FP & FN & FP & FN & FP \\
    \hline
    Assistive token-matching annotation & 2 & 1 & 1 & 4 & 5 & 4 & 0 \\
    \hline
    Assistive token-matching + disambiguation & 1 & 4 & 4 & 1 & 3 & 3 & 3 \\
    \hline
\end{tabular}}
\caption{Multi-class false positive (FP) and false negative (FN) counts for identity labeling human computation task on CivilComments}
\label{table:exp1_fp_fn_cc}
\end{table}

\subsubsection{C4 performance as control} 
\begin{figure}[htbp]
\centering
\includegraphics[width=0.9\columnwidth]{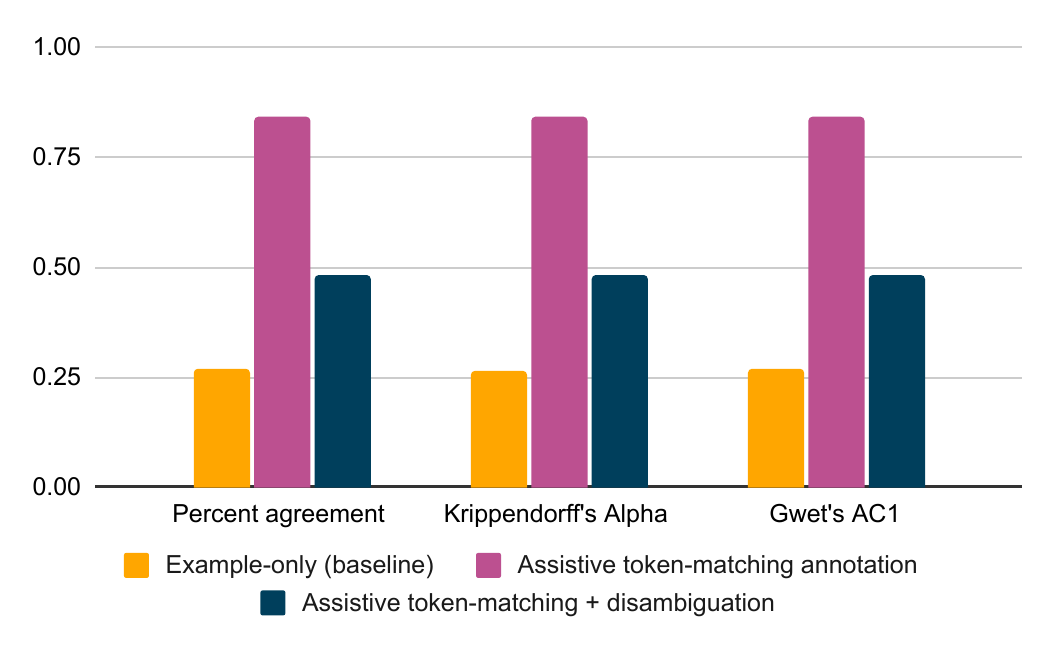}
\caption{IAR for human annotations: identity labeling on C4.}
\label{fig:hcomp_exp1_iar_c4}
\end{figure}

\begin{figure}[htbp]
\centering
\includegraphics[width=0.9\columnwidth]{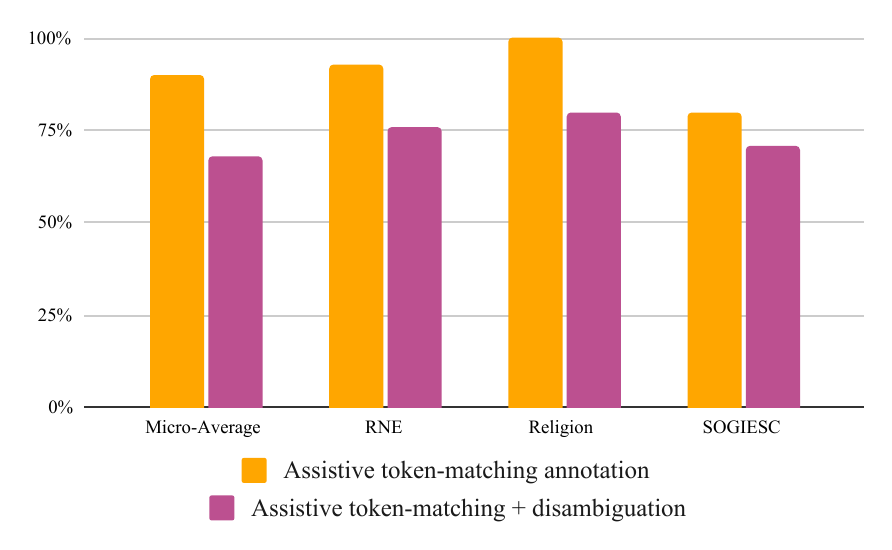}
\caption{Multi-class F1 scores for human annotations: identity labeling on C4.}
\label{fig:hcomp_exp1_f1_c4}
\end{figure}

We use C4 as control to corroborate the impact of assistive annotation for identity labeling. We run three variations of human annotation tasks, similar to CivilComments. We use the output of the example-only (no assistive annotations) task as the ground truth. Results show similar performance on both datasets. The IAR improvement (Figure~\ref{fig:hcomp_exp1_iar_c4}) of token-matching is more prominent in C4 than in CivilComments, when compared to the baseline. Similarly F1 scores (Figure~\ref{fig:hcomp_exp1_f1_c4}) are consistently better for token-matching across all groups.

\section{Fairness Applications}
\subsection{Data and models} \label{appendix_data_models}

\subsubsection{Dataset preprocessing.} We use the original splits of the CivilComments dataset \citep{JigsawUn23:online} for classifier training and evaluation. We only include toxicity labels where rater agreement is greater than 0.5. All input data is lower-cased for annotation.

\subsubsection{Model training.} All models are trained for 11 epochs with a dropout rate of 30\%, using an early stopping patience window of 3 epochs. 

\subsection{Counterfactuals} \label{appendix_counterfactuals}
\subsubsection{Similarity logic.} We use the nnlm-en-dim128\footnote{\url{https://tfhub.dev/google/nnlm-en-dim128/1}} embedding to compute similarity between terms in the lexicon. To create a counterfactual mapping we first generate a subspace of the embedding which constitutes terms for an identity group that exist in its vocabulary. To find the least similar terms, we compute the linear distance from the reflection of the term around the center of the space. The center is the average value of all vectors in the embedding subspace. 

\subsubsection{Candidate generation.} To generate counterfactuals we first annotate terms with identity groups and subgroups. We then replace all terms in a text with their corresponding counterfactuals. To address cases where identity impacts toxicity, we only generate counterfactuals for labels which are not expected to be influenced by identity i.e. identity attack greater than or equal to 0.5.

\subsection{Human computation design details} \label{appendix_hcomp_exp2_toxicity_details}
We adapt the annotation instructions from Perspective API\footnote{\url{https://github.com/conversationai/conversationai.github.io}} for our toxicity labeling task. Similar to the Perspective API process we discard NOT SURE human annotations and map TOXIC and VERY TOXIC to 1.0, and NOT TOXIC to 0.0. 

We ask the human annotators to answer the toxicity question and identity identity-based attack question (enabled only if the answer of to toxicity question was VERY TOXIC, TOXIC or HARD TO SAY). We also ask human annotators to also provide a reason for their selection.

We do not highlight any tokens or provide context for the baseline task. For the assistive-annotation tasks, we highlight the tokens and provide the context associated with them. The test questions for these tasks are created using the strategy in Appendix~\ref{appendix_hcomp_exp1_details}.

\begin{figure}[htbp]
\centering
\includegraphics[width=0.9\columnwidth]{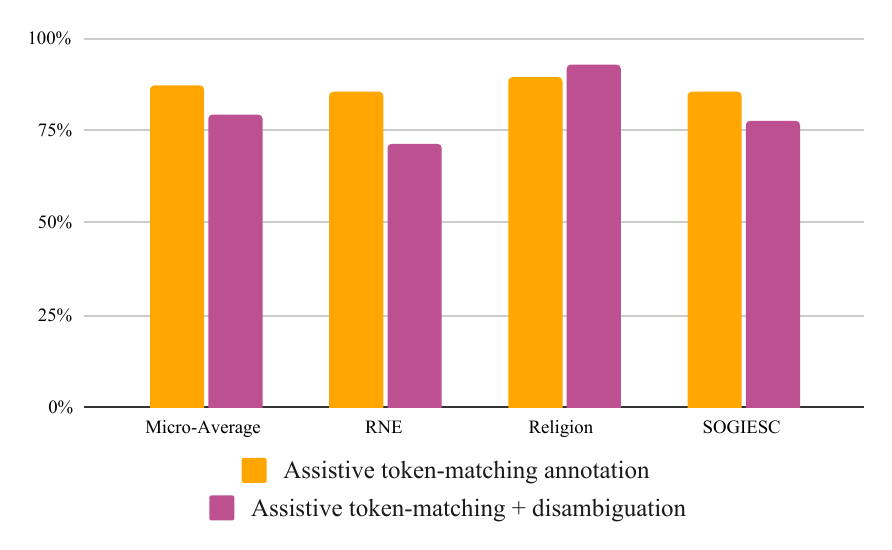}
\caption{Multi-class F1 scores for human annotations: toxicity labeling on CivilComments.}
\label{fig:hcomp_exp2_f1_cc}
\end{figure}

\begin{table}[htbp]
\resizebox{.95\columnwidth}{!}{\begin{tabular}{|p{1in}|c|c|c|c|}\hline
     & \textbf{Velocity} & \textbf{Cost} & \textbf{Ease of Job}  & \textbf{Pay} \\
    \hline
     & Judgement Time (s) & Total Judgements  & Scale: 1-5 & Scale: 1-5 \\
    \hline
    Example-only (baseline) & 33 & 11987 & 2.4 & 3.3 \\
    \hline
    Assistive Identity Group annotations & 60 & 12014 & 3.6 & 3.4 \\
    \hline
    Assistive Identity Group + Connotation annotations & 46 & 12140 & 2.5 & 3.5 \\
    \hline
\end{tabular}}
\caption{Velocity, cost and satisfaction results from human annotation tasks for toxicity labeling}
\label{table:hcomp_exp2_velocity}
\end{table}

\subsection{Human computation results}
\label{appendix_hcomp_exp2_toxicity_results}
\subsubsection{Pre-processing.} For F1 score computation, we discard NOT SURE from toxicity human annotations, and map TOXIC and VERY TOXIC to 1 and NOT TOXIC to 0 for a binary output. Similarly we discard NOT SURE from identity attack human annotations, and map YES to 1 and NO to 0. We then use F1 binary average scores to gauge the overall performance and the output of the example only (no assistive annotations) job as ground truth for this comparison. 
\subsubsection{Qualitative analysis.} The assistive IdentityGroup+Connotation task achieves the highest F1 score for both toxicity and identity-based labeling attack labels. The difference in performance is more pronounced in toxicity labeling (Figure~\ref{fig:hcomp_exp2_f1_cc}).

The human annotation task with no assistive identity context performs the best in terms of velocity, taking 45\% and 28.26\% less time than the assistive IdentityGroup and IdentityGroup+Connotation tasks, respectively (Table~\ref{table:hcomp_exp2_velocity}). Cost-wise, the baseline task is slightly better than the assistive tasks, although they all perform similarly. In the optional satisfaction survey, human annotators find the IdentityGroup+Connotation task to be easier to perform (33.33\%) and have slightly better pay compared to the baseline task.

The assistive IdentityFacet+Connotation annotation improves the IAR in human computation tasks for toxicity labeling compared to the baseline. However, the assistive IdentityFacet annotation leads to higher IAR for the ``Identity based attack'' question. This could indicate that showing Connotations might bias toxicity labels while showing IdentityGroups might bias identity-based attack labels.

Considering all the above, providing assistive identity context for task labeling should be approached carefully since it may lead to unintended bias in the labels required for model training and testing.
\end{appendices}

\end{document}